\documentclass{article}

\usepackage{microtype}
\usepackage{graphicx}
\usepackage{subfigure}
\usepackage{booktabs} % for professional tables

\usepackage{hyperref}

\usepackage{array}

\usepackage[accepted]{icml2023}

\usepackage{amsmath}
\usepackage{amssymb}
\usepackage{mathtools}
\usepackage{amsthm}
\usepackage{graphicx}
\usepackage{float}
\usepackage[capitalize,noabbrev]{cleveref}

\theoremstyle{plain}

\theoremstyle{definition}

\theoremstyle{remark}

\icmltitlerunning{On the Usefulness of Synthetic Tabular Data Generation}

\begin{document}

\twocolumn[
\icmltitle{On the Usefulness of Synthetic Tabular Data Generation}

\icmlsetsymbol{equal}{*}

\begin{icmlauthorlist}
\icmlauthor{Dionysis Manousakas}{comp}
\icmlauthor{Sergül Aydöre}{comp}
%\icmlauthor{Firstname3 Lastname3}{comp}
%\icmlauthor{Firstname4 Lastname4}{sch}
%\icmlauthor{Firstname5 Lastname5}{yyy}
%\icmlauthor{Firstname6 Lastname6}{sch,yyy,comp}
%\icmlauthor{Firstname7 Lastname7}{comp}
%\icmlauthor{}{sch}
%\icmlauthor{Firstname8 Lastname8}{sch}
%\icmlauthor{Firstname8 Lastname8}{yyy,comp}
%\icmlauthor{}{sch}
%\icmlauthor{}{sch}
\end{icmlauthorlist}

%\icmlaffiliation{yyy}{Department of XXX, University of YYY, Location, Country}
\icmlaffiliation{comp}{Amazon AWS AI/ML}
%\icmlaffiliation{sch}{School of ZZZ, Institute of WWW, Location, Country}

\icmlcorrespondingauthor{Dionysis Manousakas}{\href{mailto:dionman@amazon.co.uk}{dionman@amazon.co.uk}
}
%\icmlcorrespondingauthor{Firstname2 Lastname2}{first2.last2@www.uk}

\icmlkeywords{Machine Learning, ICML}

\vskip 0.3in
]

\printAffiliationsAndNotice{}  % leave blank if no need to mention equal contribution

\begin{abstract}

Despite recent advances in synthetic data generation, the scientific community still lacks a unified consensus on its usefulness. It is commonly believed that synthetic data can be used for both data exchange and boosting machine learning (ML) training. Privacy-preserving synthetic data generation can accelerate data exchange for downstream tasks, but there is not enough evidence to show how or why synthetic data can boost ML training. In this study, we benchmarked ML performance using synthetic tabular data for four use cases: \emph{data sharing}, \emph{data augmentation}, \emph{class balancing}, and \emph{data summarization}. We observed marginal improvements for the balancing use case on some datasets. However, we conclude that there is not enough evidence to claim that synthetic tabular data is useful for ML training.
\end{abstract}

\section{Introduction}
\label{sec:introduction}

Generative modeling is a powerful technique that can be used to create synthetic data that mimics the properties of a given population of real data. This can be a valuable tool for machine learning applications, as it can help to improve predictive ability and fairness.

When working with scarce or imbalanced datasets ~\citep{sos, ai23}, generative modeling can be used to augment the data by creating synthetic data points that fill in the gaps. This can help to improve the performance of machine learning models, as they will have more data to train on.

In addition to improving predictive ability, generative modeling can also be used to improve the fairness of machine learning models. This is because synthetic data can be created to be more representative of the population as a whole, which can help to reduce bias in the models. For example, a study by~\citet{decaf} found that augmenting a dataset with synthetic data can help to improve the fairness of a machine learning model for predicting recidivism rates.

In this study, our focus is tabular data, which poses unique challenges for generative modeling, including:
\begin{itemize}
\item Lack of local structure: Tabular data does not have the same local structure as image, audio, or text data. This makes it difficult for generative models to learn the relationships between different features.
\item Co-existence of categorical and numerical features: Tabular data often contains a mix of categorical and numerical features. This can make it difficult for generative models to learn a joint distribution over all of the features.
\item Feature missingness: Tabular data often contains missing values. This can make it difficult for generative models to learn a complete distribution over all of the features.
\item High scarcity: Tabular data is often scarce. This means that there is not enough data to train a generative model.
\end{itemize}

As a result of these challenges, advances in generative models for tabular data have lagged behind advances in image, audio, and text data ~\citep{stasy, tabddpm}. In order to address these challenges, researchers have developed tailored synthetic data quality metrics. These metrics can be used to evaluate the quality of synthetic tabular data, and to help ensure that the data is suitable for use in machine learning applications.

In this study, we conduct a range of experiments to evaluate the applicability of synthetically generated data for generic downstream tabular learning tasks. We focus on four axes of practical interest:

\begin{itemize}

    \item \emph{Data sharing / Synthetic data quality}:     
We evaluate the downstream performance of models trained on synthetic data and compare it to the performance of models trained on the original data distribution. For synthetic data to be useful, machine learning development performed on synthetic data should lead to the same conclusions as if it were carried out on the real data. The train-on-synthetic and test-on-real (TS-TR) paradigm is the default experiment for evaluating the usefulness of generative modeling methods for data sharing.
Here are some additional details about the TS-TR paradigm:
\begin{itemize}
    \item In the TS-TR paradigm, a generative model is trained on a dataset of real data.
    \item The generative model is then used to generate synthetic data.
    \item A machine learning model is trained on the synthetic data.
    \item The performance of the machine learning model is evaluated on a hold-out dataset of real data.
\end{itemize}
The TS-TR paradigm is a useful tool for evaluating the usefulness of generative modeling methods for data sharing. However, it is important to note that the TS-TR paradigm does not provide formal privacy guarantees. If privacy is a concern, additional noise may need to be added to the synthetic data as in \citet{vietri2022private}.

    \item \emph{Summarization}: In this setting, we evaluate whether we can use synthetic data to represent the original distribution with fewer samples. To do this, we replace a given data population with a controllable, smaller number of synthetic data points sampled from a generative model that has been trained on the full real dataset. We then use the synthetic data for downstream learning tasks. In contrast to the \emph{data sharing} study outlined above, where we fixed the volume of synthetic data to equal the real data used for the generative model training, in the TS-TR setting of this experiment, we focus on the trade-off between the number of synthetic data points and the achievable downstream ML efficiency.
    
    \item \emph{Augmentation}: To address the issue of data scarcity, original training data can be augmented with synthetic data to create a richer training dataset. In this setting, we augment real data populations with synthetically generated data before performing downstream learning tasks. We then optimize the end learner on the augmented training dataset and evaluate it on real test data. We call this approach train-on-augmented, test-on-real (TA-TR). 
    
    \item \emph{Class balancing}: For classification tasks, we can sample synthetic data conditionally from the minority categories of a given dataset to reduce or eliminate class imbalance prior to downstream training. We then use the balanced version of the data as a training dataset for the downstream classifier (TA-TR). This setting differs from the previous one in that we sample synthetic data conditionally on the minority class.

\end{itemize}

 For the synthetic data generation, we consider two generative modelling approaches (cf.~\cref{subsec:downstream-models}): (i) deep generative models~(DGMs) trained from scratch using a real data population at hand, and (ii) fine-tuned versions of pre-trained large language models~(LLMs) that incorporate prior information beyond the real data at hand. These experiments aim to unveil if we can achieve an equivalent or better representation of the training data distribution for general, downstream model-agnostic learning purposes, via leveraging synthetic samples from generative models. We also examine whether the synthetic samples have the capacity to efficiently compress statistical information from the training data population without further guidance, allowing relying on a smaller sample size for downstream tasks, without statistical quality loss. This could unlock the potential of accelerating downstream learning tasks. 

With the exception of the synthetic data quality evaluation, the rest of our experiments have a smaller footprint in the literature. Studies in deep generative model based data balancing have reported only weak improvements to lightweight baselines in the tabular domain~\citep{elor}, with only recently proposed specialised generative architectures being able to more confidently outperform SMOTE~\citep{sos}. In addition, we are unaware of experiments with generative models addressing tabular data summarization, while an augmentation setting similar to ours has been considered in~\citet{goggle} reporting no significant downstream boosting. 

\begin{table*}[!h]
    \centering
    \caption{Dataset statistics and task details.}
    \label{tab:dataset_stats}
    \begin{tabular}{lccccccc}
        \toprule
        Dataset & Domain & \#Samples & \#Num & \#Cat & Task & \#Classes & Imbalance $\left(\frac{\text{\#Majority pts.}}{\text{\#Minority pts.}}\right)$ \\
        \midrule
        adult & Social & 32,561 & 6 & 8 & Classification & 2 & 3.15 \\
        churn & Marketing & 954 & 2 & 4 & Classification & 2 & 3.26 \\
        sick & Medical & 3,772 & 7 & 22 & Classification & 2 & 11.96 \\
        heloc & Financial & 9,871 & 21 & 2 & Classification & 2 & 1.1 \\
        california & Real Estate & 20,640 & 8 & 0 & Regression & - & -\\
        \bottomrule
    \end{tabular}
\end{table*}

\section{Benchmarking}
\label{sec:benchmarking}

Below we give an overview of the components involved in our benchmarking setup. These include a set of generative models~(\cref{subsec:generative-models}) that enable us to sample synthetic datapoints w/ and w/o conditioning, a set of downstream learners~(\cref{subsec:downstream-models}) which correspond to the end learning task for the data distribution at hand, and some public access tabular datasets~(\cref{subsec:datasets}) that we use for our evaluation. We conclude this section with further details on our experimental setup~(\cref{subsec:experimental-setup}).

\begin{figure*}[!t]
    \begin{minipage}[t]{0.245\textwidth}
        \centering
        \includegraphics[width=\linewidth]{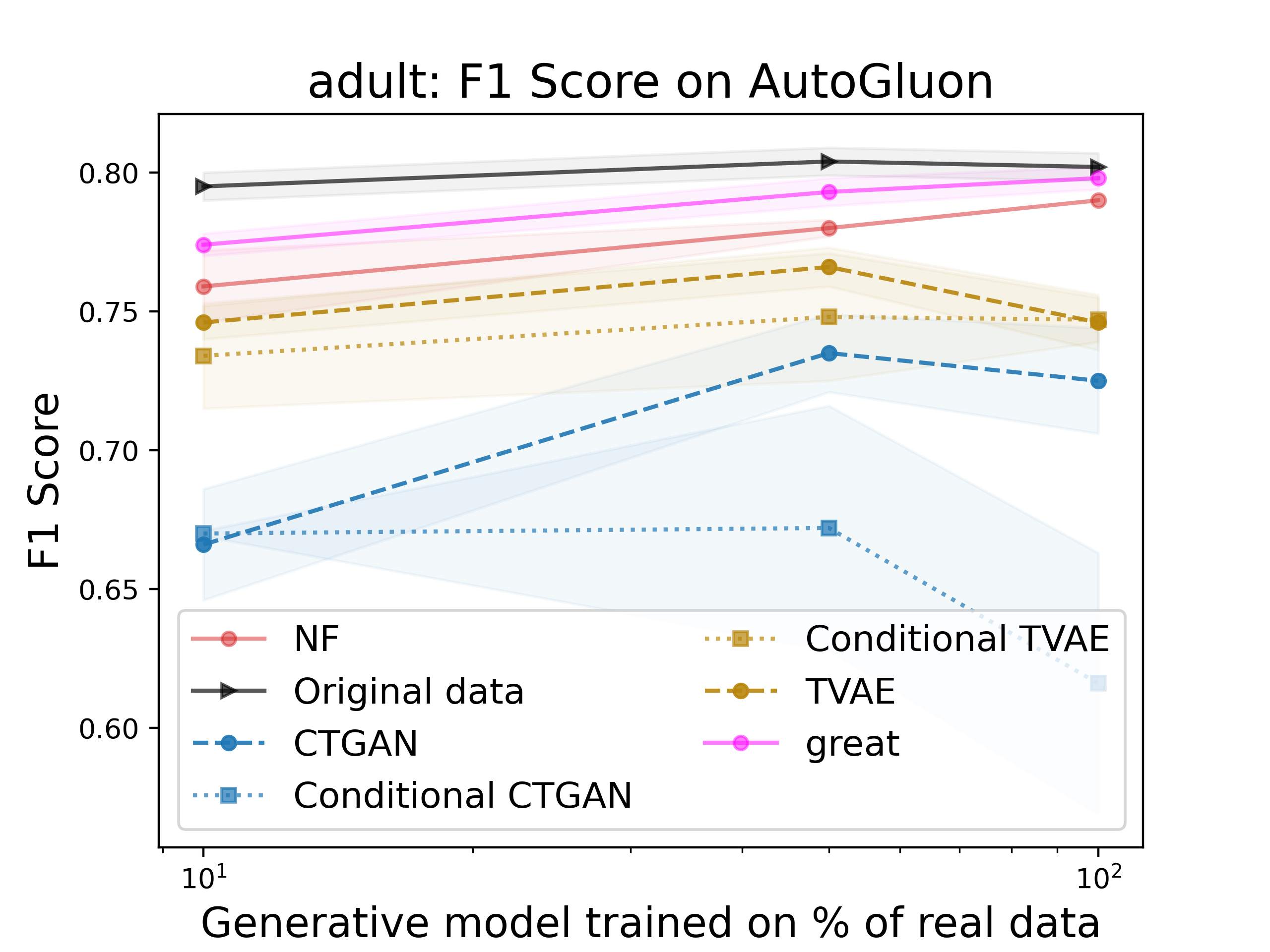}
    \end{minipage}%
    \hfill
    \begin{minipage}[t]{0.245\textwidth}
        \centering
        \includegraphics[width=\linewidth]{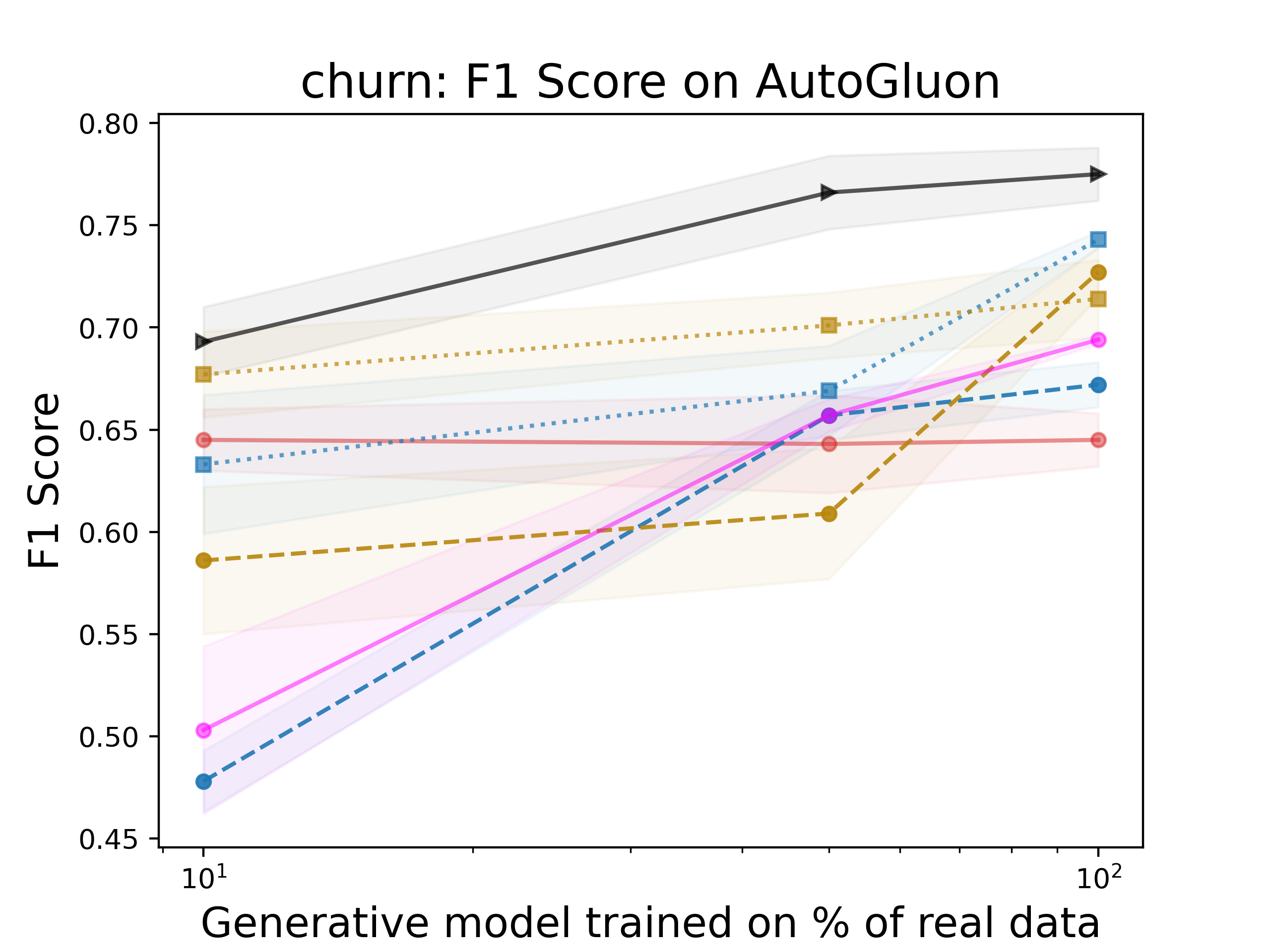}
    \end{minipage}
    \hfill
    \begin{minipage}[t]{0.245\textwidth}
        \centering
        \includegraphics[width=\linewidth]{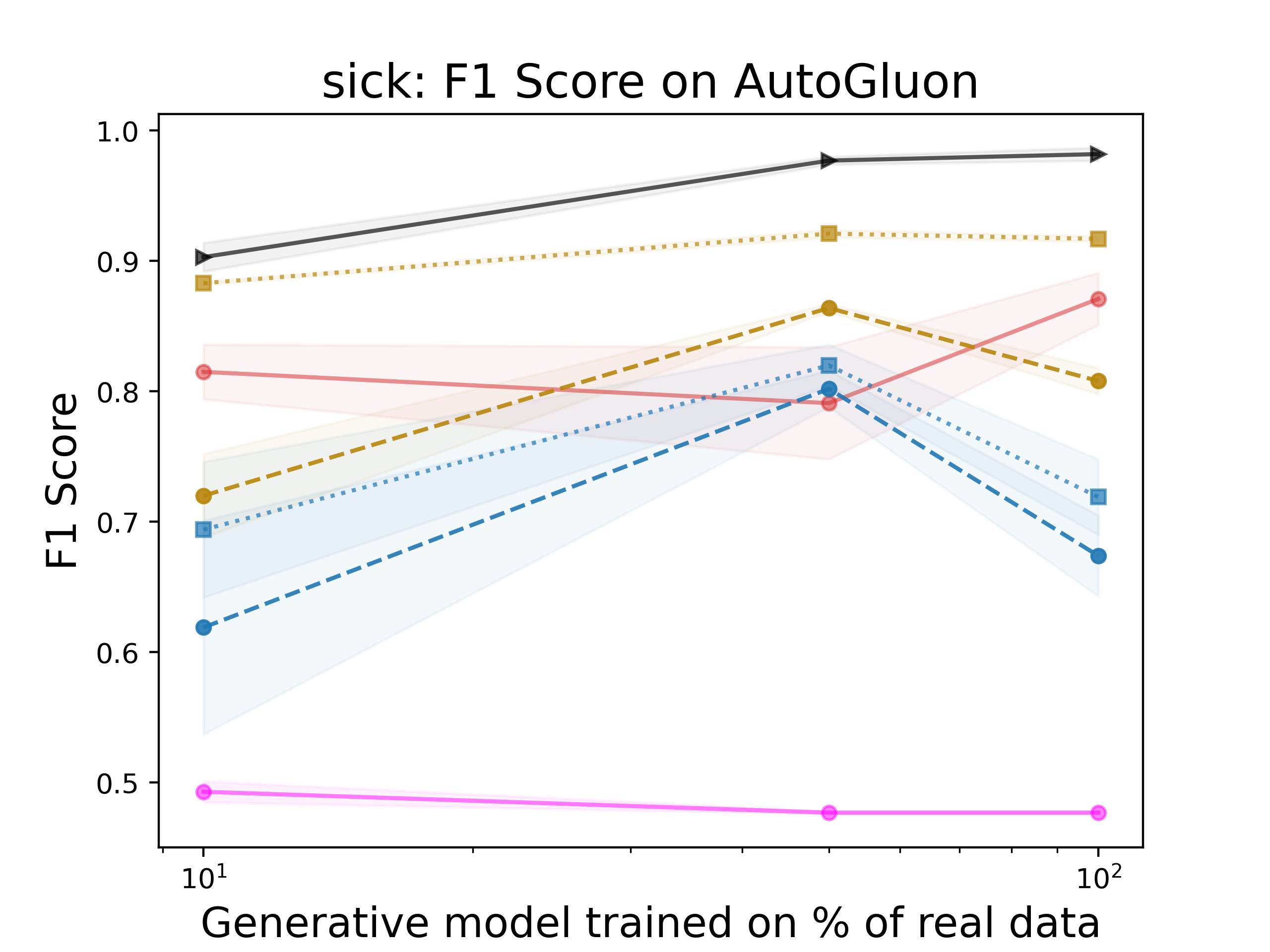}
    \end{minipage}
    \hfill
    \begin{minipage}[t]{0.245\textwidth}
        \centering
        \includegraphics[width=\linewidth]{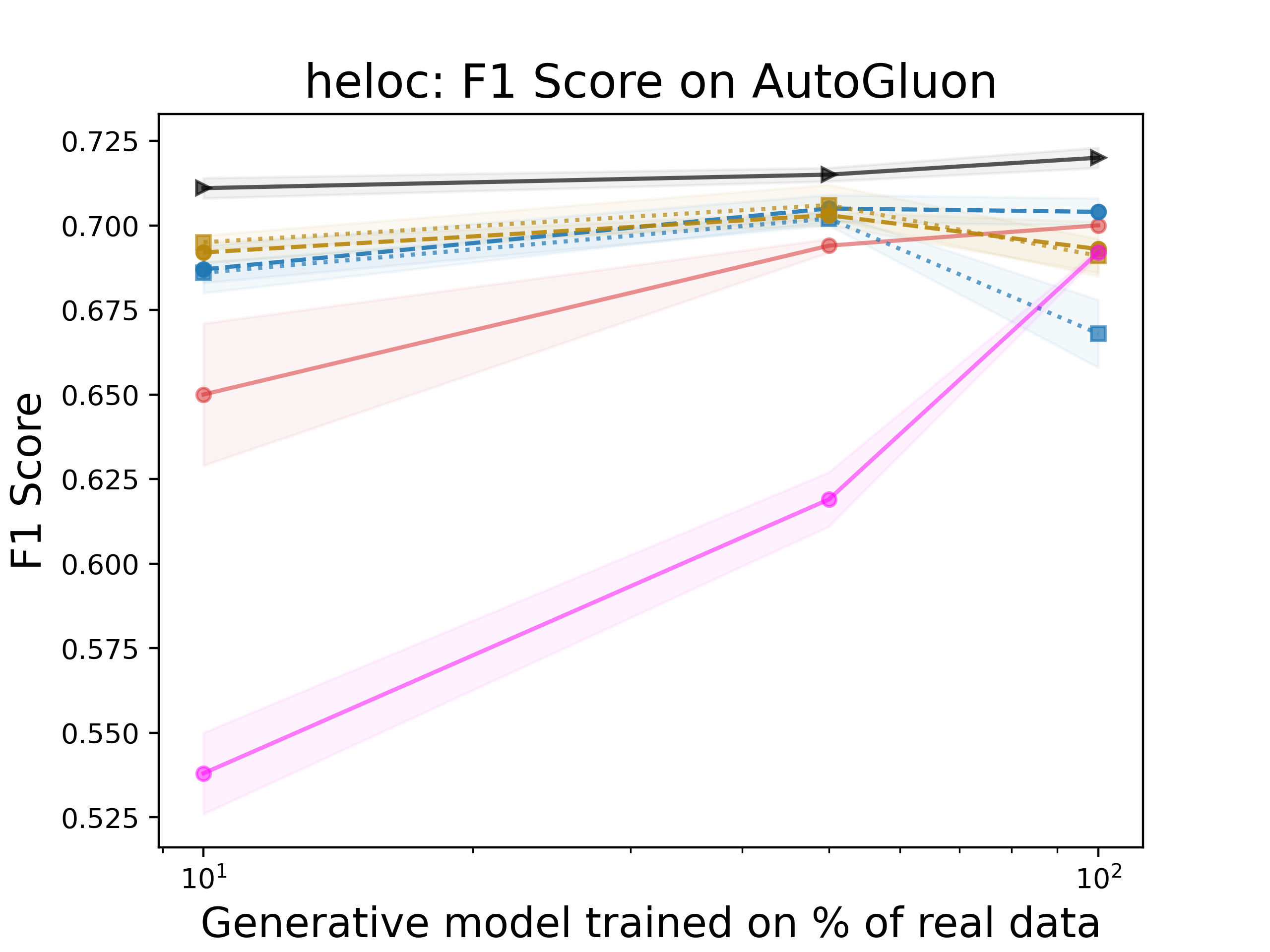}
    \end{minipage}
    \caption{(\emph{Synthetic data quality experiment}) Predictive performance of a downstream AutoGluon classifier, which uses synthetic training data sampled from generative models that have been trained on increasing \emph{subsets} of the real data. The number of synthetic training datapoints increases along the $x$-axis, matching the number of real data used for training the generative model.}
    \label{fig:synthetic-classification}
\end{figure*}

\begin{figure*}[!t]
    \begin{minipage}[t]{0.245\textwidth}
        \centering
        \includegraphics[width=\linewidth]{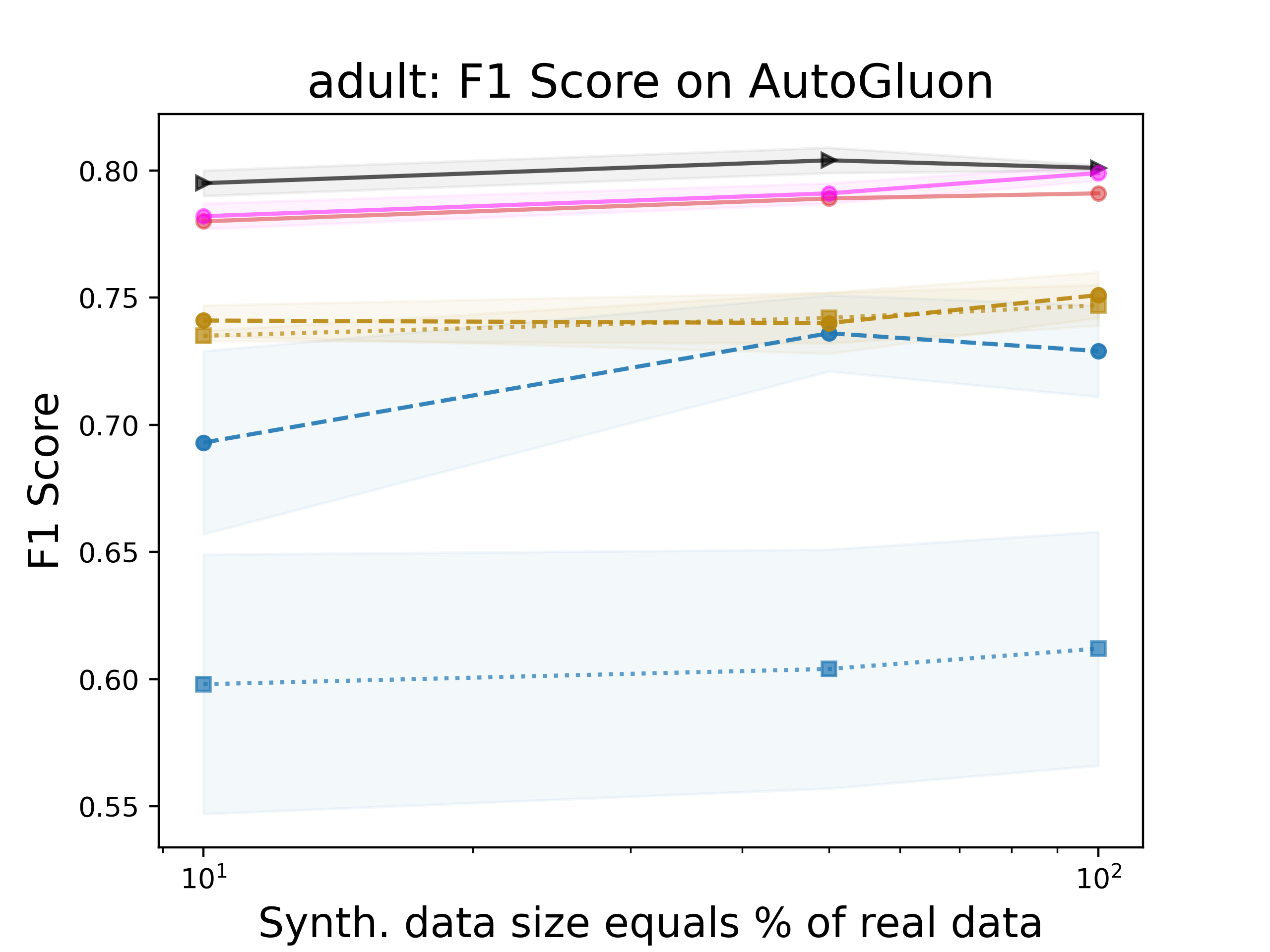}
    \end{minipage}%
    \hfill
    \begin{minipage}[t]{0.245\textwidth}
        \centering
        \includegraphics[width=\linewidth]{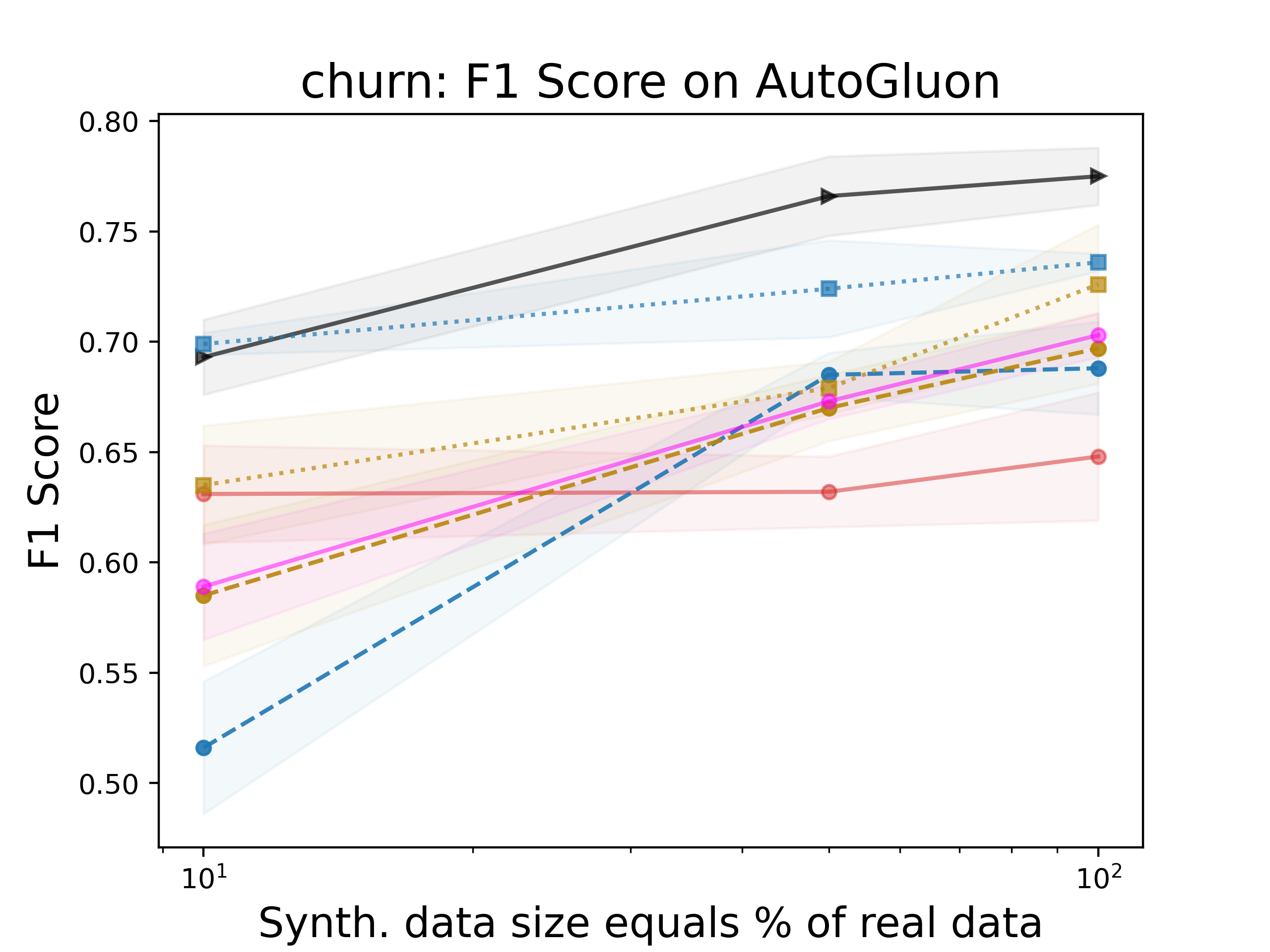}
    \end{minipage}
    \hfill
    \begin{minipage}[t]{0.245\textwidth}
        \centering
        \includegraphics[width=\linewidth]{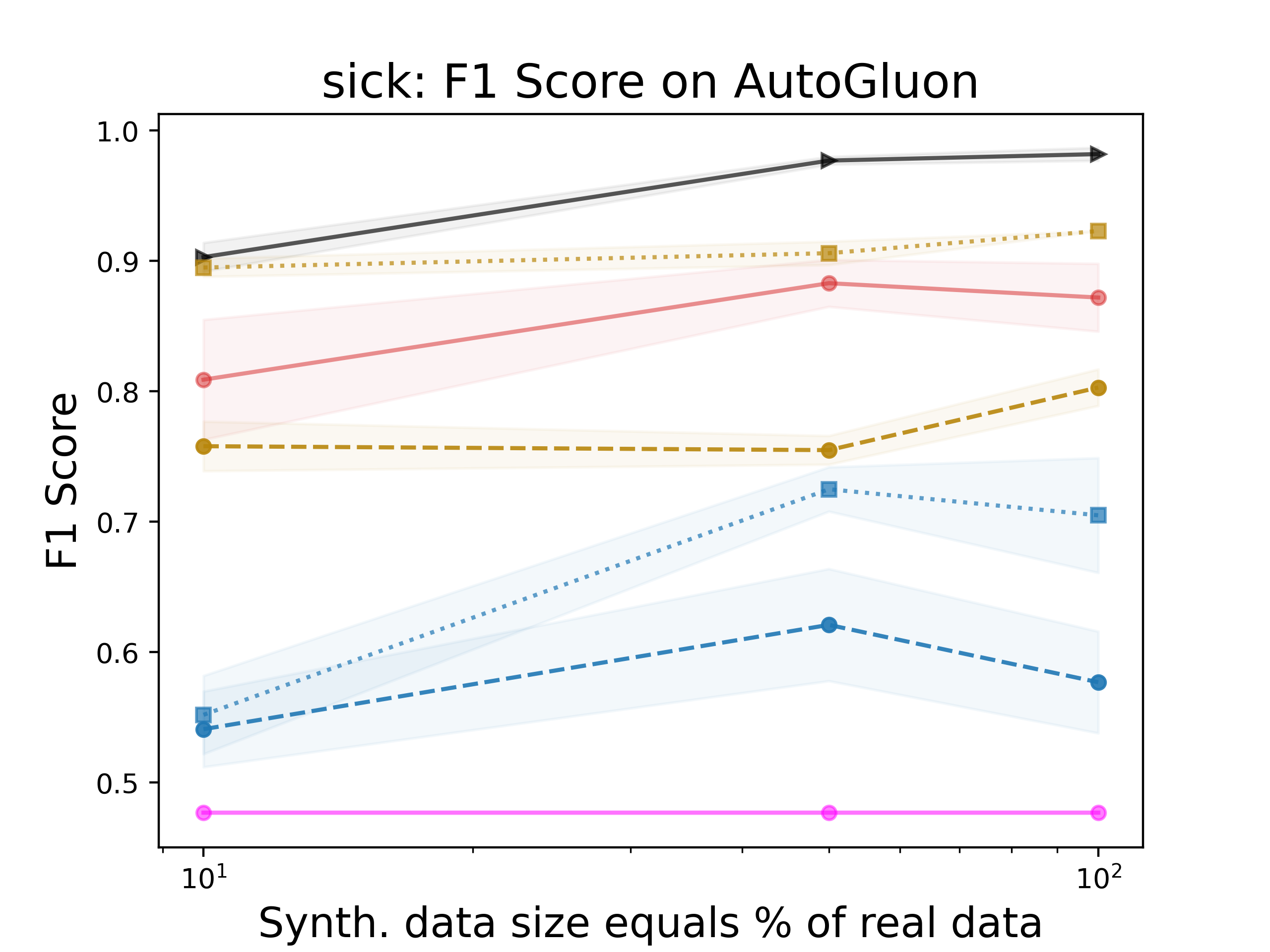}
    \end{minipage}
    \hfill
    \begin{minipage}[t]{0.245\textwidth}
        \centering
        \includegraphics[width=\linewidth]{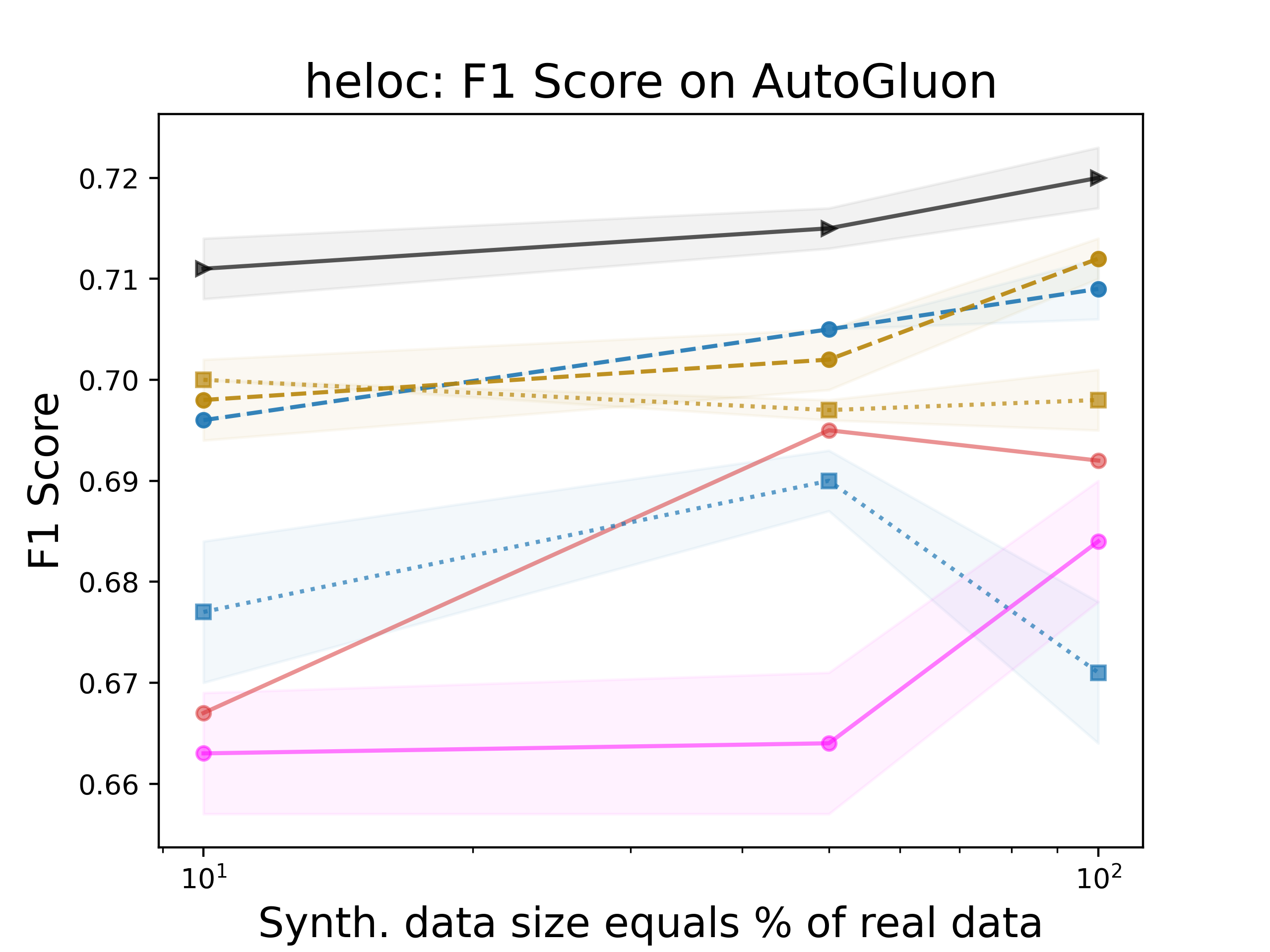}
    \end{minipage}
    \caption{(\emph{Data summarization experiment}) Predictive performance of a downstream AutoGluon classifier which uses synthetic training data sampled from a generative model that has been trained on the \emph{full} real dataset. The number of synthetic training data increases along the $x$-axis.}
    \label{fig:class-summarization}
\end{figure*}

\subsection{Generative models}
\label{subsec:generative-models}

We include the following state-of-the-art tabular data generation methods in our experimentation.

\paragraph{Conditional Tabular GANs (CTGANs)}\citep{ctgan}. An adaptation of the original GAN architecture~\citep{gans} with two main modifications for handling tabular datasets: (i) an elaborate normalization scheme for multi-modal features, and (ii) a training-by-sampling paradigm for better learning of under-represented categories. Training-by-sampling means that a random column is being selected and conditioned on during each training iteration of the GAN. We overload this notation, referring with the term \textbf{Conditional CTGAN} to a CTGAN model that is trained to learn the conditional distribution $p(x|y)$ instead, where $y$ is a target variable.\footnote{as opposed to modeling $p(x, y)$ which is the original learning goal of CTGAN} As in~\citet{conditional-gan}, training in the case of Conditional CTGAN differs from the original training-by-sampling paradigm in that the same conditional column $y$ will be used throughout training. The latter allows conditional sampling on the target variable for the class balancing use case, as at sampling time it enables passing explicitly the information of the desired class $y$ and using the CTGAN to sample $x \sim p(x|y)$, to jointly generate a synthetic sample $(x,y)$.  Moreover, it makes use of factorization of the joint that is based on the downstream learning task.

\paragraph{Tabular Variational AutoEncoders (TVAEs)}\citep{ctgan}. An adaptation of the original VAE architecture~\citep{vae} sharing the pre-processing steps of CTGAN. Similarly to CTGANs, we also consider \textbf{Conditional TVAEs} that model directly $p(x|y)$, where $y$ is the corresponding target variable.

\paragraph{Normalizing Flows (NFs)}. A normalizing flow~\citep{papamak} is a generative model that transforms a simple distribution to a more complex one by applying a series of invertible transformations. These transformations gradually distort the simple distribution to match the target distribution of the data. By optimizing the parameters of the flow through maximum likelihood estimation, the model learns to generate new samples from the learned distribution and estimate the likelihoods of observed data points accurately. In our tabular learning experiments, we use the method proposed in~\citet{nfs} with standard one-hot encoding for categorical features. We only use normalizing flows to learn the data joint probability, hence omit this model from the balancing use case that requires conditional sampling.

\paragraph{Fine-tuned pre-trained Large Language Models (GReaT)}. We use the method proposed in~\citet{great}. Training consists of (i) converting the tabular datapoints into sentences via a simple parsing scheme, and (ii) fine-tuning an existing large-language model on the data. We rely on tuning the \emph{distill-GPT2} model~\citep{distilbert} on our tabular datapoints at hand for each benchmark (introduced in the original paper as ~\emph{distil-GReaT}). We adopt the key design choices selected in~\citet{great}, e.g. a perturbations scheme on the extracted sentences. GReaT uses auto-regressive neural networks implementing a transformer decoder-only architecture~\citep{transformers}, which along with the perturbation scheme, enable arbitrary sampling on the datapoint features (including conditioning on the target variable, which is used in the balance use case experiments).

\paragraph{SMOTE}~\citep{smote} A non-parametric method addressing class imbalance in datasets. It works by generating synthetic samples for the minority class by interpolating between existing minority class instances. The algorithm selects a minority class sample and finds its $k$ nearest neighbors. It then randomly selects one or more neighbors and creates new synthetic samples along the line segments joining the original sample and its neighbors. This process helps to increase the representation of the minority class, balancing the class distribution and improving the model ability to learn from the minority class instances.

\paragraph{Original} A baseline that subsamples true datapoints uniformly at random in the absence of augmentation. Random sampling controls for the varying sample size effects, capturing the expected performance on a specified volume of real training data w/o interventions via generative models.

For our experiments with DGMs we relied on the implementations of the \texttt{synthcity} package~\citep{synthcity23}, for GReaT we used the original implementation of the authors \texttt{be\_great} \citep{great}, while for SMOTE we used the implementation of the model for mixed numeric and categorical data from~\citet{imbalanced-learn}.

\subsection{Downstream models and metrics}
\label{subsec:downstream-models}

We are interested in measuring the extra downstream utility induced via generative models for broad practical use cases. Hence, to make our evaluation agnostic to a specific category of downstream ML models, for \emph{classification} tasks we consider two powerful tabular classifiers, namely XGBoost (XGB)~\citep{xgboost}, and AutoGluon~\citep{agtabular}. For \emph{regression} tasks, we evaluate against a linear regression model (LinReg), a multilayer perceptron (MLPRegressor) and AutoGluon. We present the metrics obtained on AutoGluon in the main part, deferring the remaining models to~\cref{app:additional-results}. We keep F1-Score and RMSE as the evaluation metrics respectively for classification and regression.

\subsection{Datasets}
\label{subsec:datasets}

We give a short overview of the datasets used in our experiments in~\cref{tab:dataset_stats}. We rely on publicly available datasets (cf. ~\cref{tab:dataset_urls}) comprised of 4 datasets for classification experiments, and 1 for regression, with sizes varying from less than 1k datapoints to more than 30k.

\subsection{Experimental setup}
\label{subsec:experimental-setup}

We use a fixed split of training and testing datapoints with a $80/20$ percent scheme throughout the entire experiment.
For each experiment we repeat the training/fine-tuning of the corresponding generative model across 3 independent trials, and present the obtained mean and standard error of the computed metrics.
We make use of a fixed set of training hyperparameters across datasets for each generative model. We apply hyperpameter optimization (HPO) independently per trial for each combination of use case and model over downstream learning. For HPO we use asynchronous random search. Please refer to~\cref{app:reproducibility} for details on the considered hyperparameter search spaces. As the number of nearest neighbors used to construct artificial minority samples for SMOTE we use $k=5$ throughout. In the case of GReaT we fine-tuned the distil-GPT model for a total of 200 epochs for each of our datasets, and used the default hyperparameters for training and sampling. For DGMs we trained up to a maximum of 5,000 epochs using early-stopping with a patience of 500 epochs without improvement on a validation score.

\begin{figure*}[!t]
    \begin{minipage}[t]{0.245\textwidth}
        \centering
        \includegraphics[width=\linewidth]{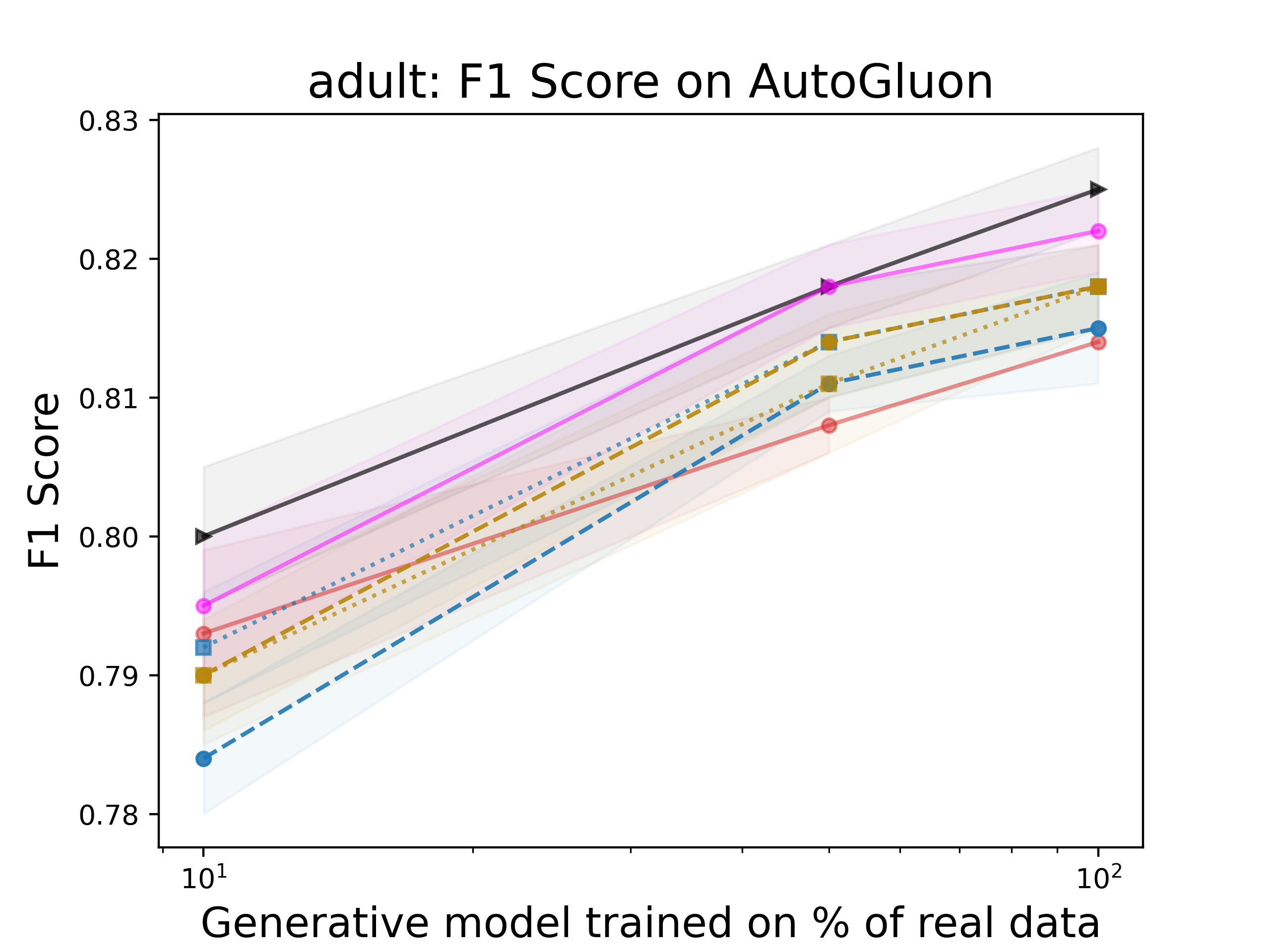}
    \end{minipage}%
    \hfill
    \begin{minipage}[t]{0.245\textwidth}
        \centering
        \includegraphics[width=\linewidth]{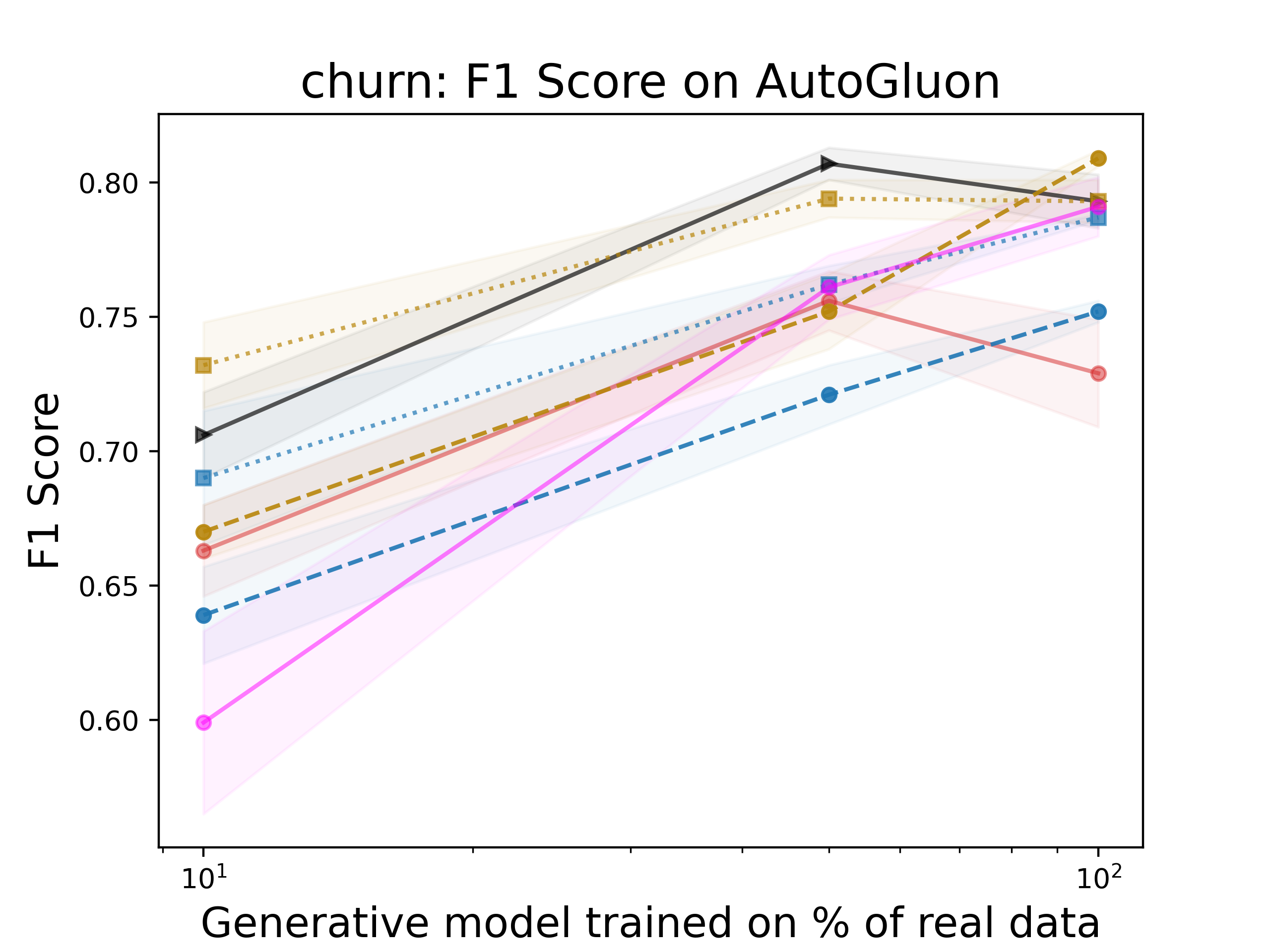}
    \end{minipage}
    \hfill
    \begin{minipage}[t]{0.245\textwidth}
        \centering
        \includegraphics[width=\linewidth]{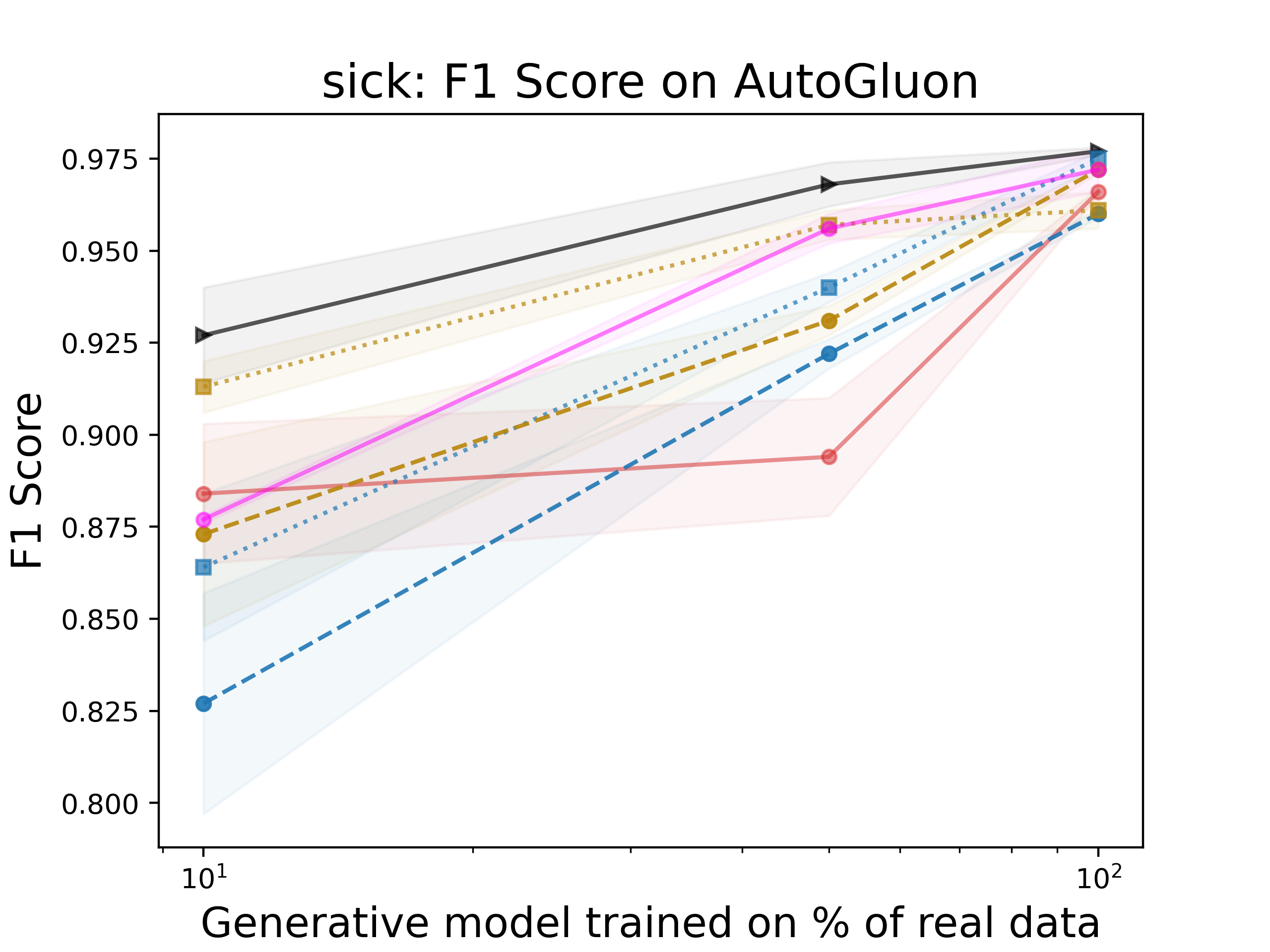}
    \end{minipage}
    \hfill
    \begin{minipage}[t]{0.245\textwidth}
        \centering
        \includegraphics[width=\linewidth]{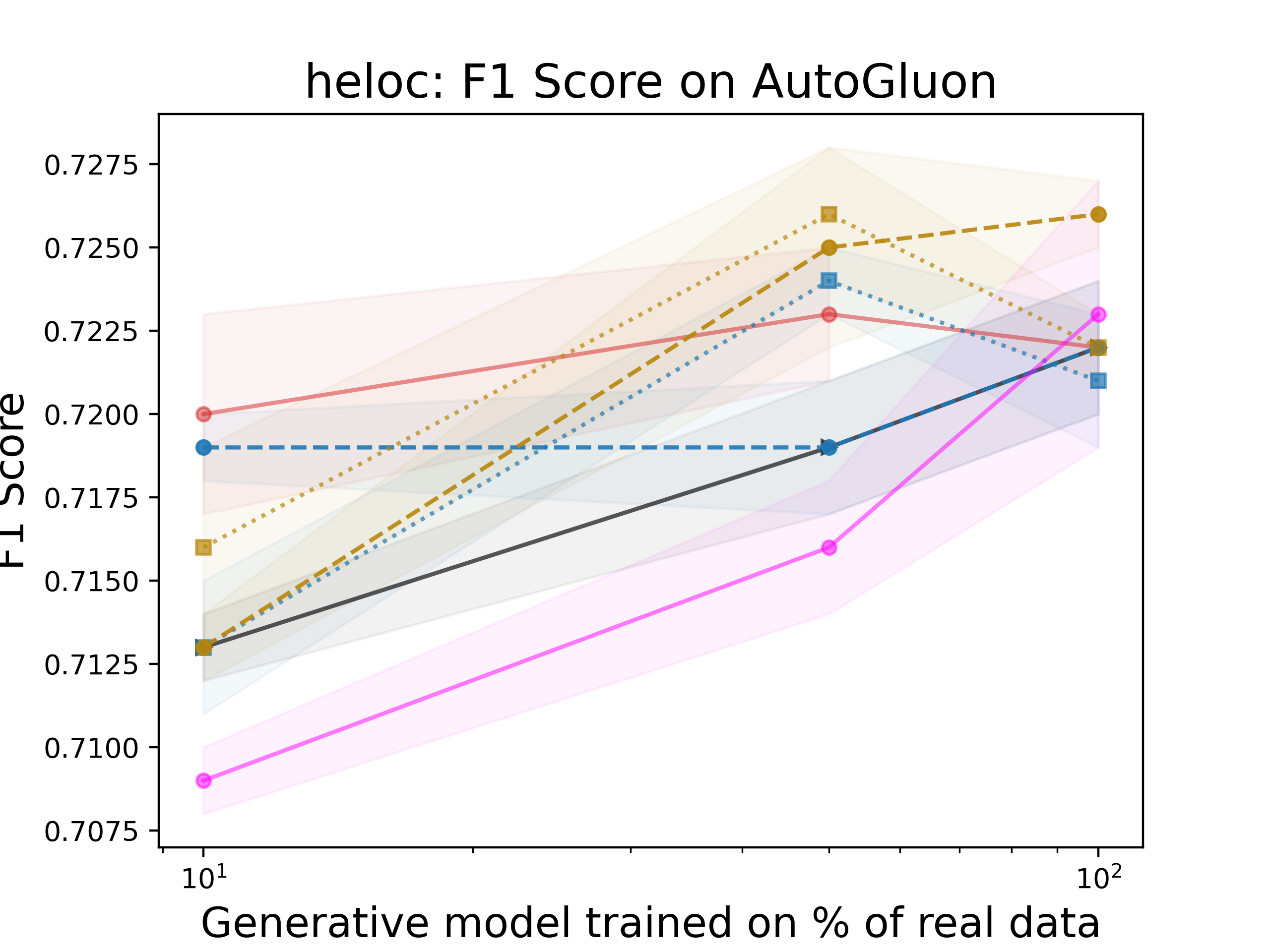}
    \end{minipage}
    \caption{(\emph{Data augmentation experiment}) Predictive performance of a downstream AutoGluon classifier which uses increasing subsets of real data \emph{augmented} with equal-sized volumes of synthetic data.}
    \label{fig:class-augmentation}
\end{figure*}

\begin{figure*}
\begin{minipage}[t]{0.325\textwidth}
        \centering
        \includegraphics[width=\linewidth]{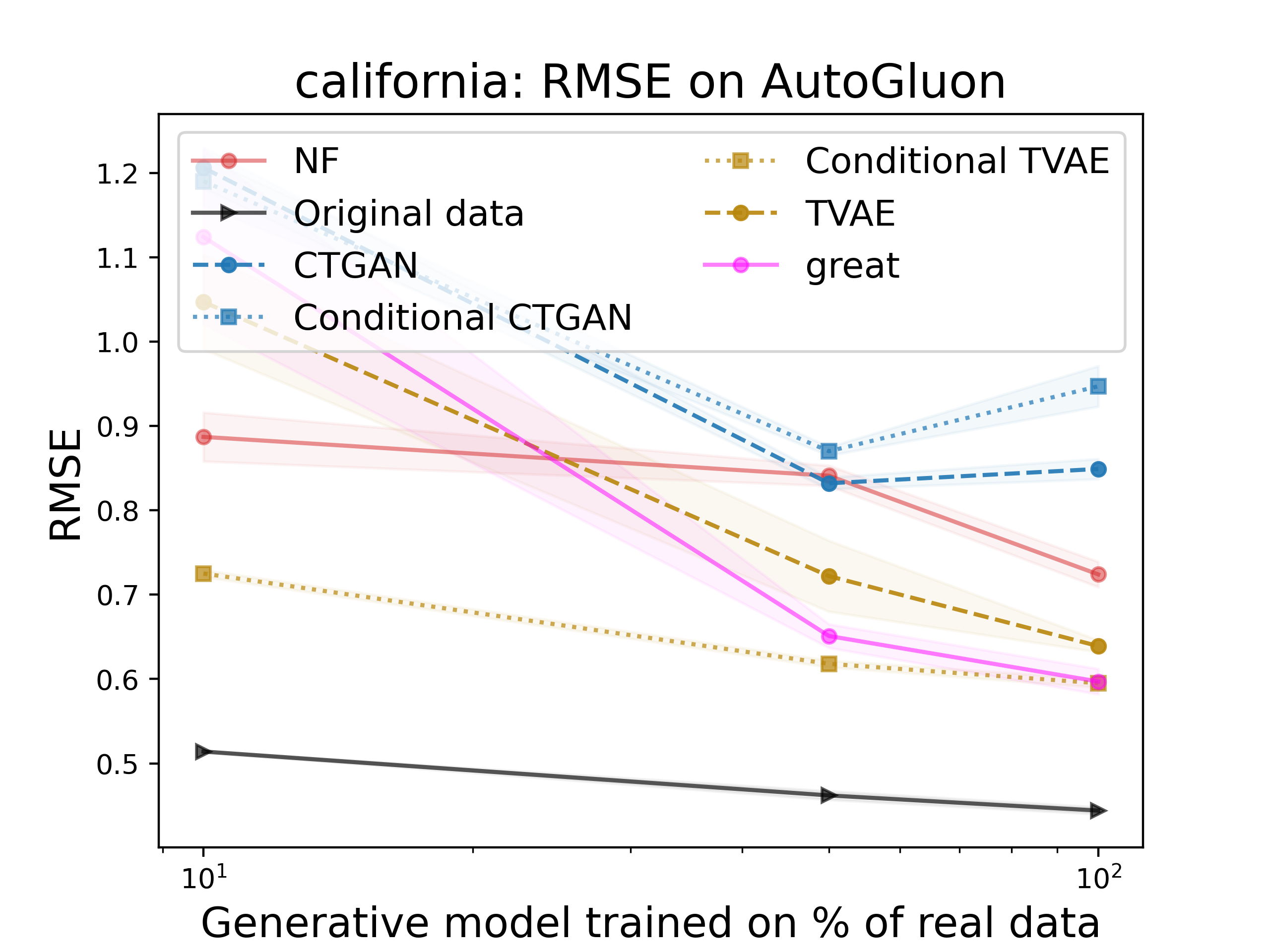}
        \caption{\emph{Synthetic data quality} experiment for regression}
    \label{fig:regression-synthesize}
    \end{minipage}
    \hfill 
\begin{minipage}[t]{0.325\textwidth}
        \centering
        \includegraphics[width=\linewidth]{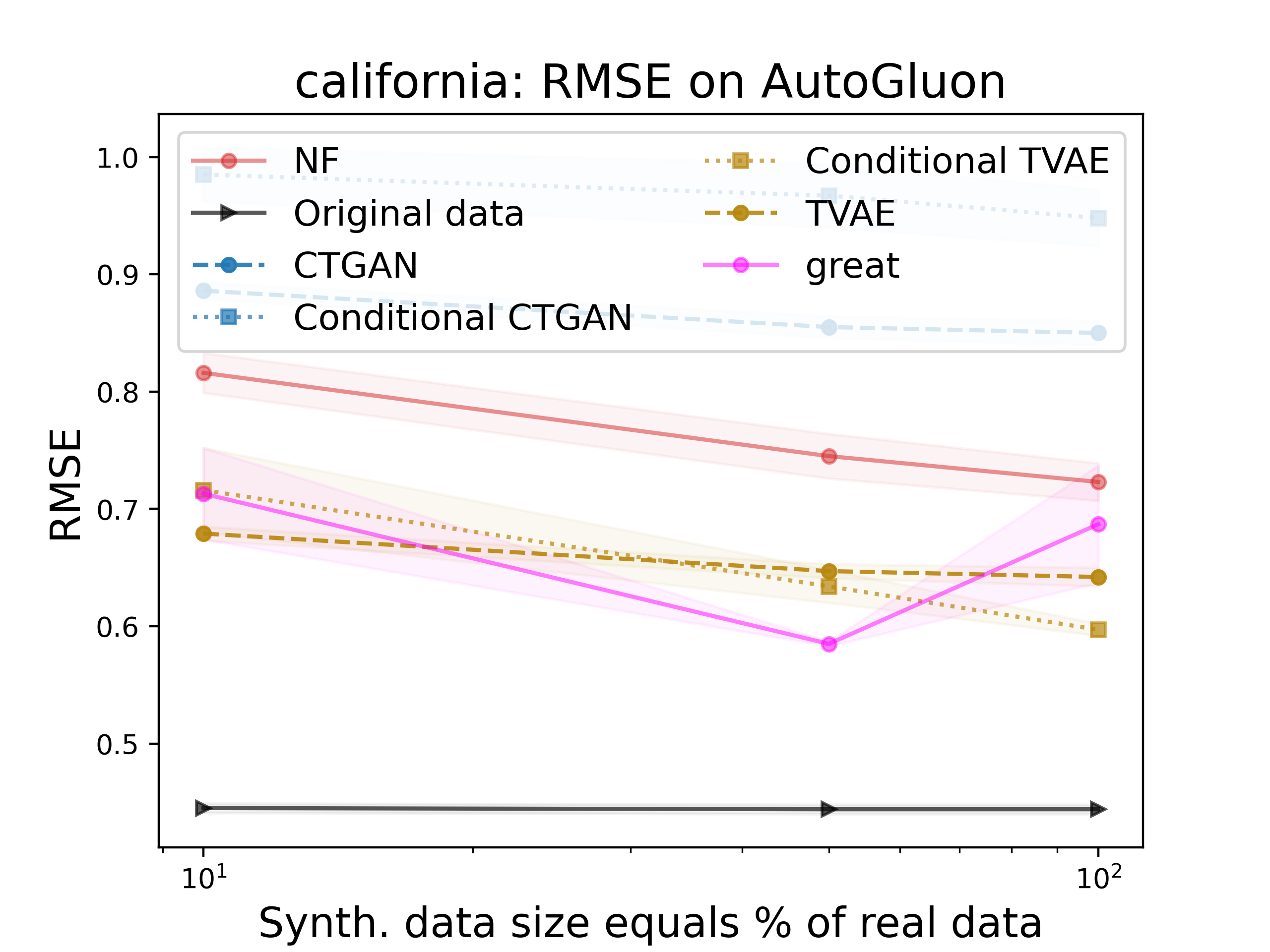}
        \caption{\emph{Data summarization} experiment for regression.}
    \label{fig:regression-summarization}
    \end{minipage}
    \hfill
    \begin{minipage}[t]{0.325\textwidth}
        \centering
        \includegraphics[width=\linewidth]{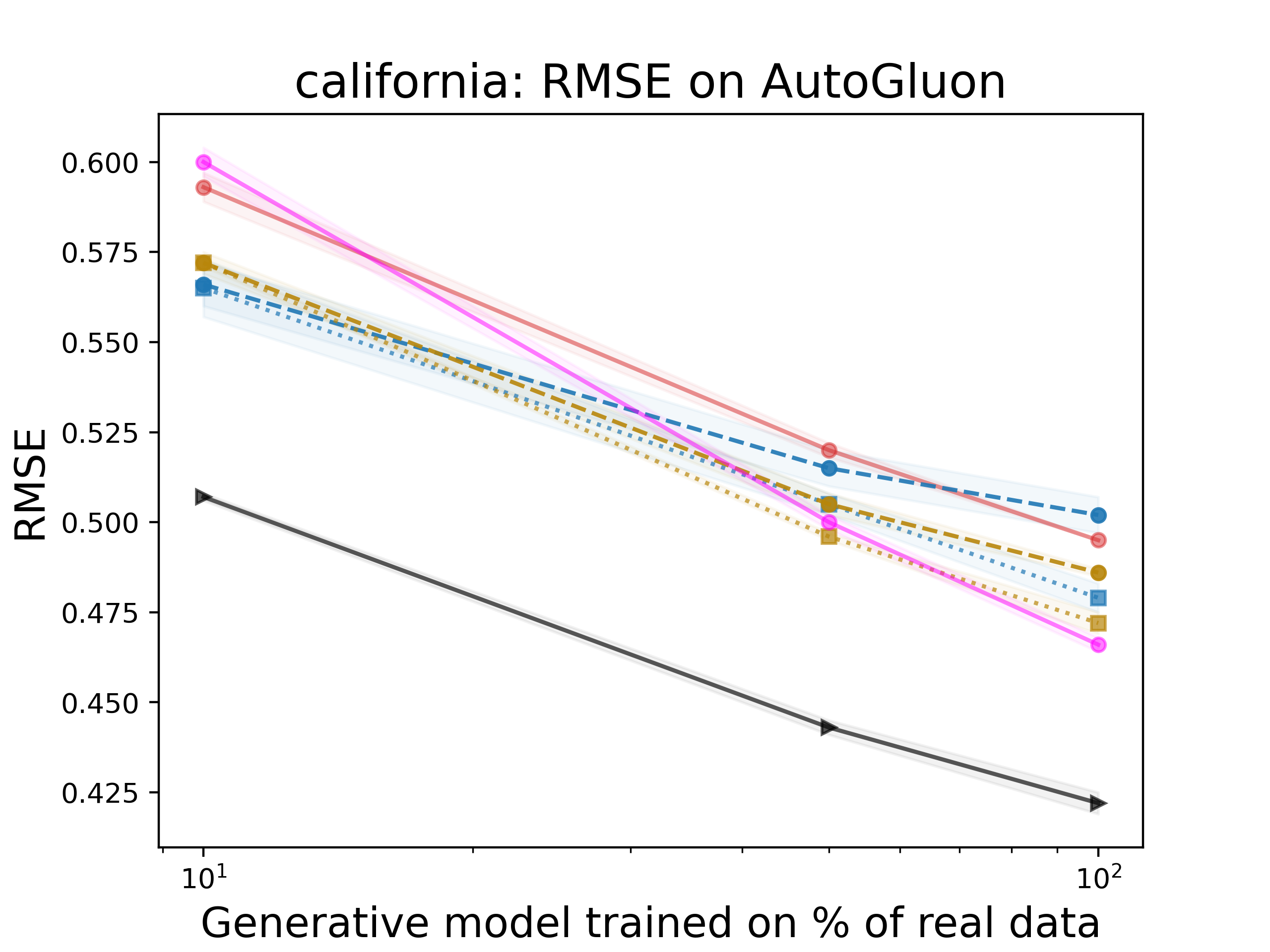}
        \caption{\emph{Data augmentation} experiment for regression.}
    \label{fig:regression-augmentation}
    \end{minipage}
\end{figure*}

\section{Results}

\label{sec:results}

%\lipsum[1-15] % Generates Lorem Ipsum text for three pages

\subsection{Quality of Synthetic Data}
\label{subsec:sd-quality}
In this part of the experiment, we aim to evaluate the statistical quality of the synthetic samples for downstream learning tasks. For this purpose, we remove the original datapoints prior to training the downstream model, and replace them with an \emph{equal volume of synthetic datapoints}. To capture the effects of varying sample sizes of the true data, we report predictive performance for the described TS-TR setup, considering generative models trained on increasing fractions of the original training population ($10$, $50$ and $100\%$). The obtained downstream performance is shown in~\cref{fig:synthetic-classification,fig:regression-synthesize}. As a general remark, training on the available original data outperforms training on the synthetic samples for all tested generative models. Regarding the latter, we observed that the relative ordering of the downstream efficacy of the compared DGMs does not seem to change much across the varying dataset size regimes, while the gap between synthetic and real data is closing as we use more datapoints to train the generative model. When trained on the full dataset, conditional TVAE maintains the highest rank among DGMs across the datasets. For large datasets with interpretable feature semantics (e.g. adult, see~\cref{tab:feature-names}) we eventually obtain higher quality using the pre-trained LLM. In the case of datasets with higher dimensionality, long contexts of specialised terms or abbreviated feature names, such as sick and heloc, GReaT seems to require fine-tuning on more training examples in order to eventually provide synthetic data quality at the same levels with DGMs. 

\begin{table*}[]
\centering
\begin{tabular}{@{}lccccc@{}}
\toprule
 &
  Original data &
  SMOTE &
  \begin{tabular}[c]{@{}c@{}}Conditional \\ CTGAN\end{tabular} &
  \begin{tabular}[c]{@{}c@{}}Conditional \\ TVAE\end{tabular} &
  GReaT \\ \midrule
 &
  \multicolumn{5}{c}{F1 Score} \\
adult &
  0.802±0.009 &
  0.809±0.008 &
  0.808±0.009 &
  \textbf{0.812±0.008} &
  0.805±0.006 \\
churn &
  0.779±0.023 &
  \textbf{0.808±0.011} &
  0.741±0.003 &
  0.771±0.02 &
  0.777±0.02 \\
sick &
  \textbf{0.982±0.009} &
  0.974±0.007 &
  0.917±0.013 &
  0.921±0.015 &
  0.977±0.003 \\
heloc &
  0.721±0.004 &
  0.717±0.003 &
  \textbf{0.722±0.003} &
  0.721±0.002 &
  0.72±0.004 \\
 &
  \multicolumn{5}{c}{ROC AUC} \\
adult &
  0.908±0.005 &
  \multicolumn{1}{l}{0.906±0.004} &
  \multicolumn{1}{l}{0.91±0.004} &
  \multicolumn{1}{l}{\textbf{0.913±0.003}} &
  \multicolumn{1}{l}{0.908±0.003} \\
churn &
  \textbf{0.934±0.008} &
  \multicolumn{1}{l}{0.917±0.01} &
  \multicolumn{1}{l}{0.899±0.005} &
  \multicolumn{1}{l}{0.913±0.013} &
  \multicolumn{1}{l}{0.915±0.008} \\
sick &
  \textbf{0.999±0.0} &
  \multicolumn{1}{l}{\textbf{0.999±0.001}} &
  \multicolumn{1}{l}{0.994±0.001} &
  \multicolumn{1}{l}{\textbf{0.999±0.001}} &
  \multicolumn{1}{l}{\textbf{0.999±0.0}} \\
heloc &
  0.787±0.005 &
  \multicolumn{1}{l}{0.784±0.005} &
  \multicolumn{1}{l}{0.786±0.005} &
  \multicolumn{1}{l}{\textbf{0.791±0.006}} &
  \multicolumn{1}{l}{0.787±0.005} \\ \bottomrule
\end{tabular}
\caption{(\emph{Data balancing experiment}) Predictive metricsevaluated on a downstream AutoGluon classifier that uses an rebalanced augmented version of the dataset via samples from the generative models.}
\label{tab:balance-tab}
\end{table*}

\subsection{Summarization}
\label{subsec:summarize}
Here we investigate if sampling from generative models trained on the full original data distribution inherently enables more efficient compression of the statistical information of the original data for the downstream task, compared to extracting random subsets of real datapoints. As in the previous section, we replace the original datapoints entirely with synthetic samples with the help of a generator; however now the generator has been trained on the full real data, and we use volumes of \emph{synthetic data with varying size} corresponding to 10, 50 and 100\% of the true data population. As we observe in~\cref{fig:class-summarization,fig:regression-summarization}, the achievable downstream quality is generally lower compared to equal volumes of real datapoints, and marginally higher compared to the corresponding points in~\cref{fig:synthetic-classification,fig:regression-synthesize}, as now the generative models have seen more training examples before producing the same amount of synthetic samples. This hints that sampling from these generative models does not significantly reduce the redundancy in the resulting population to lower levels compared to the original data;  ML efficiency seems to be following slightly flatter curves for the synthetic data training sets, and saturates at the same levels with the plots of the synthetic data quality experiment~(\cref{subsec:sd-quality}) for synthetic populations with the same volume as the real data.

\subsection{Augmentation}
\label{subsec:augmentation}
In this experiment, we train our generative models on increasing subsets of the original data population, comprising $10$, $50$ and $100\%$ of the full data, and complement with equal volumes of synthetic data, sampled from a trained generative model that has seen only the subset. Finally, \emph{we train the downstream classifier on the augmented volume of real and synthetic datapoints}. This setting aims to simulate situations with data scarcity, and provide empirical evidence addressing whether augmenting with generative models can benefit downstream predictive accuracy. We present the ML  efficacy achieved with this augmentation strategy for classification and regression tasks in~\cref{fig:class-augmentation,fig:regression-augmentation}. We observe that augmentations via generative models were not capable to consistently improve the performance that we got when constraining the downstream training data only among the original datapoints.

\subsection{Class balancing}
\label{subsec:balancing}
In our last scenario, we consider the effects of eliminating the imbalance ratio of the original training datapoints via synthetic data from generative models. We follow the same setup with the augmentation use case, developing a TA-TR experiment for evalution. However here, we restrict our analysis on the conditional DGMs that have been trained to model $p(x|y)$, and \emph{extract synthetic samples corresponding to the minority category} by setting $y$ to the minority category. We also sample minority datapoints from GReaT starting the autoregressive generation via setting the feature $y$ to the value of the minority class. Hence, we can restore the balance among the categories of each dataset. Subsequently, we provide the downstream model with the balanced version of the data as the new richer training dataset. As observed in~\cref{tab:balance-tab} the generative models are capable to only marginally outperform the original data and the SMOTE baseline in 2 of the 4 considered datasets. 

% Please add the following required packages to your document preamble:
% \usepackage{booktabs}

%\section{Related work}
%\label{sec:related-work}

\section{Conclusions}
\label{sec:conclusions}

We investigated the effects of generative model-based data augmentation strategies on downstream learning tasks for commonly used tabular benchmarks. We experimented with data imbalance and scarcity learning scenarios. We deliberately excluded privacy and fairness considerations from this paper to focus on the usefulness of ML training. We found that, despite ongoing advances in tabular generative modeling, popular state-of-the-art methods are not yet able to consistently outperform the ML efficacy of using only the original training datapoints for practical use cases, such as data summarization, augmentation, and learning in imbalanced datasets. We believe that including experiments that simulate real-world situations where practitioners might need synthetic data is often missing from the literature, and should be seen as a necessary prerequisite for improving tabular synthetic data generation methods. As future work, we will focus on integrating extensive HPO, adding downstream task-informed regularization terms over training of the generative models, as well as using pre-trained LLMs that are more relevant to the domain of the training data.

% Acknowledgements should only appear in the accepted version.
% \section*{Acknowledgements}

\clearpage
\newpage
\bibliography{main}
\bibliographystyle{icml2023}

%%%%%%%%%%%%%%%%%%%%%%%%%%%%%%%%%%%%%%%%%%%%%%%%%%%%%%%%%%%%%%%%%%%%%%%%%%%%%%%
%%%%%%%%%%%%%%%%%%%%%%%%%%%%%%%%%%%%%%%%%%%%%%%%%%%%%%%%%%%%%%%%%%%%%%%%%%%%%%%
% APPENDIX
%%%%%%%%%%%%%%%%%%%%%%%%%%%%%%%%%%%%%%%%%%%%%%%%%%%%%%%%%%%%%%%%%%%%%%%%%%%%%%%
%%%%%%%%%%%%%%%%%%%%%%%%%%%%%%%%%%%%%%%%%%%%%%%%%%%%%%%%%%%%%%%%%%%%%%%%%%%%%%%
\newpage
\appendix

\onecolumn

\section{Dataset details}
\label{app:data}

In this section we include some further details on the datasets that we used for our experiments. In~\cref{tab:feature-names} we outline the names of the dataset features used for our generative model training, indicating in bold the target variable for the downstream discrimative learning task. Note that, with the exception of GReaT, all generative methods are agnostic to the semantics of the features (only require the column type to apply the right encoding and preprocessing of the corresponding feature). On the other hand, for GReaT feature semantics directly interact with synthetic data quality, as the datapoints will be understood by the corresponding language model as a sentence of the form "\texttt{feature\_name\_0} \emph{is} \texttt{feature\_value\_0}, \texttt{feature\_name\_1} \emph{is} \texttt{feature\_value\_1}, \ldots", adopting the pre-trained word embeddings for the features.

In~\cref{tab:dataset_urls} we disclose the original sources of the datasets. 

% Please add the following required packages to your document preamble:
% \usepackage{booktabs}
\begin{table*}[]
\centering
\caption{Feature names of the datasets.}
\label{tab:feature-names}
\begin{tabular}{@{}l||l@{}}
\toprule
adult &
  \begin{tabular}[c]{@{}l@{}}age, workclass, fnlwgt, education, education-num,  marital-status, occupation, relationship, race, sex, \\ capital-gain, capital-loss, hours-per-week, native-country, \textbf{income}\end{tabular} \\
\midrule
churn &
  \begin{tabular}[c]{@{}l@{}}Age, Frequent Flyer, Annual Income, Class of user, Number of times Services Opted during recent years, \\ Company account of user synced To Social Media, Whether the customer book lodgings / Hotels using company services, \\ \textbf{Target}\end{tabular} \\
\midrule
sick &
  \begin{tabular}[c]{@{}l@{}}age, sex, on thyroxine, query on thyroxine, on antithyroid medication, sick, pregnant, thyroid surgery, I131 treatment, \\ query hypothyroid,  query hyperthyroid, lithium, goitre, tumor, hypopituitary, psych, TSH measured, TSH, \\ T3 measured, T3, TT4 measured, TT4, T4U measured, T4U,  FTI measured, FTI, TBG measured, \\ referral source, \textbf{binaryClass}\end{tabular} \\
\midrule
heloc &
  \begin{tabular}[c]{@{}l@{}}External Risk Estimate, Months Since Oldest Trade Open, Months Since Most Recent Trade Open, Average Months in File, \\ Number of Satisfactory Trades, Number of Trades 60+ Days Past Due - Public Records or Derogatory Public Records, \\ Number of Trades 90+ Days Past Due - Public Records or Derogatory Public Records, \\ Percentage of Trades Never Delinquent,  Months Since Most Recent Delinquency, \\ Maximum Delinquency 2 Public Records in Last 12 Months, Maximum Delinquency Ever, Number of Total Trades,  \\ Number of Trades Open in Last 12 Months,  Percentage of Installment Trades, \\ Months Since Most Recent Inquiry excluding 7 days,  Number of Inquiries in Last 6 Months,  \\ Number of Inquiries in Last 6 Months excluding 7 days, Net Fraction Revolving Burden, Net Fraction Installment Burden, \\ Number of Revolving Trades with Balance,  Number of Installment Trades with Balance, \\ Number of Bank/National Trades with High Utilization,  Percentage of Trades with Balance, \\ \textbf{Risk Performance}\end{tabular} \\
\midrule
california &
  \begin{tabular}[c]{@{}l@{}}MedInc, HouseAge, AveRooms, AveBedrms, Population, AveOccup, Latitude, Longitude, \textbf{target}\end{tabular} \\
\bottomrule
\end{tabular}
\end{table*}

\begin{table}[ht]
    \centering
    \caption{URLs for real-world datasets of the study.}
    \label{tab:dataset_urls}
    \begin{tabular}{ll}
        \toprule
        Dataset & URL \\
        \midrule
        adult~\citep{adult_data, uci_data} & \small{\url{https://archive.ics.uci.edu/ml/datasets/Adult/}} \\
        churn & \tiny{\url{https://www.kaggle.com/datasets/tejashvi14/tour-travels-customer-churn-prediction}} \\
        sick~\citep{sick_data, uci_data} & \tiny{\url{https://www.openml.org/search?type=data&sort=runs&id=38&status=active}} \\
        heloc & \tiny{\url{https://www.kaggle.com/datasets/averkiyoliabev/home-equity-line-of-creditheloc}} \\
        california & \tiny{\url{https://www.kaggle.com/datasets/camnugent/california-housing-prices}} \\
        \bottomrule
    \end{tabular}
\end{table}

\section{Additional results}
\label{app:additional-results}

\begin{figure*}[!t]
    \begin{minipage}[t]{0.245\textwidth}
        \centering
        \includegraphics[width=\linewidth]{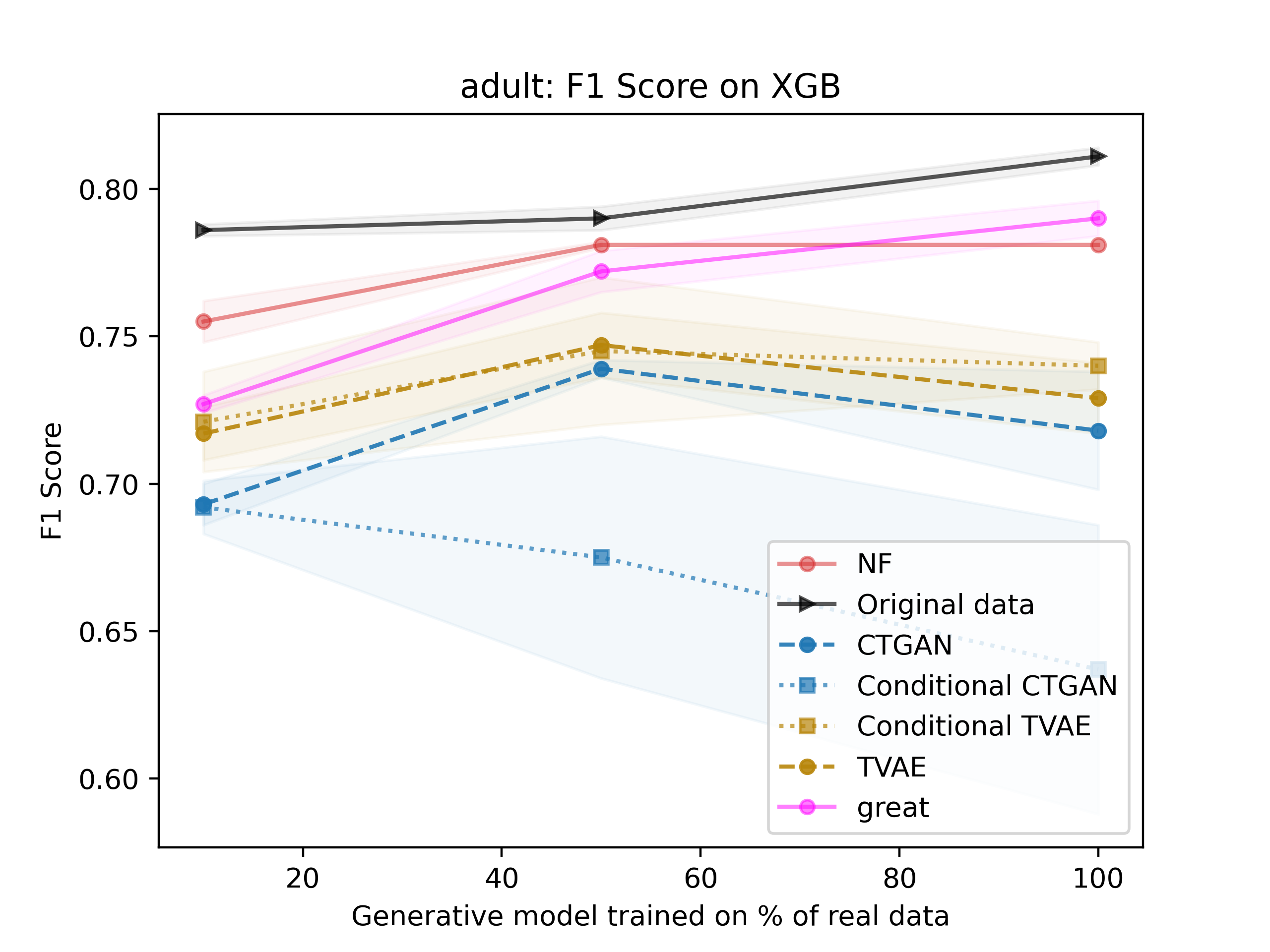}
    \end{minipage}%
    \hfill
    \begin{minipage}[t]{0.245\textwidth}
        \centering
        \includegraphics[width=\linewidth]{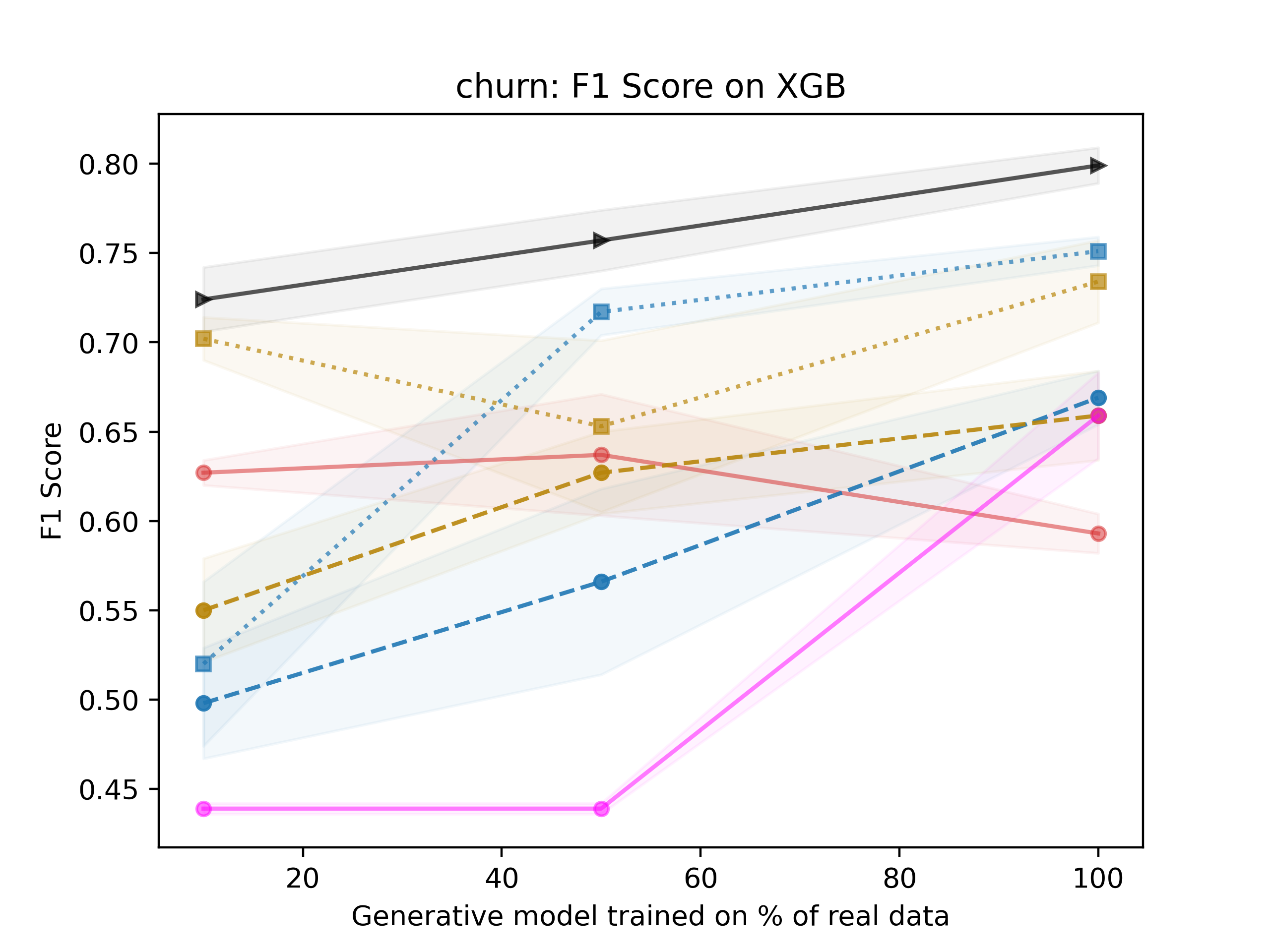}
    \end{minipage}
    \hfill
    \begin{minipage}[t]{0.245\textwidth}
        \centering
        \includegraphics[width=\linewidth]{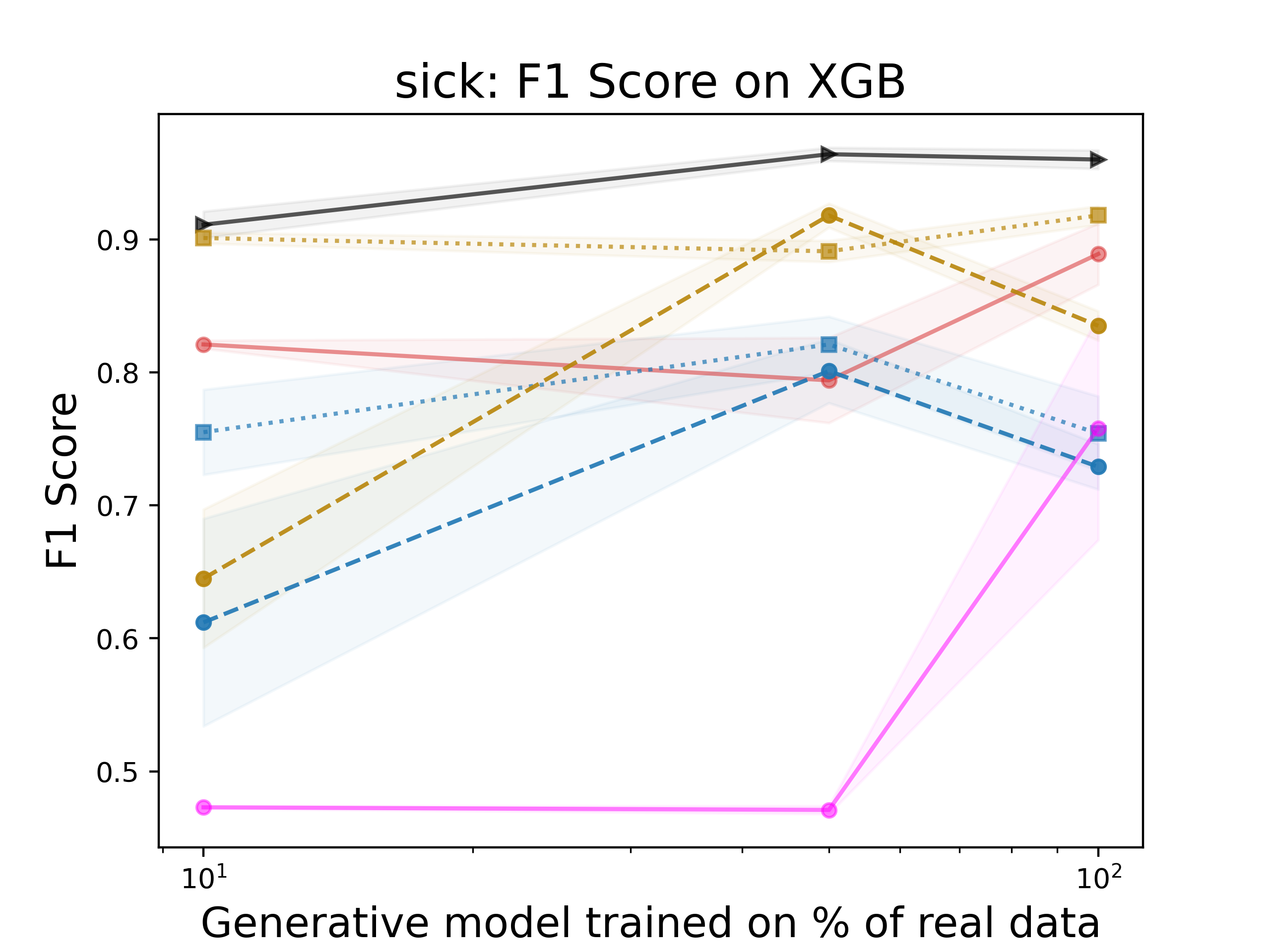}
    \end{minipage}
    \hfill
    \begin{minipage}[t]{0.245\textwidth}
        \centering
        \includegraphics[width=\linewidth]{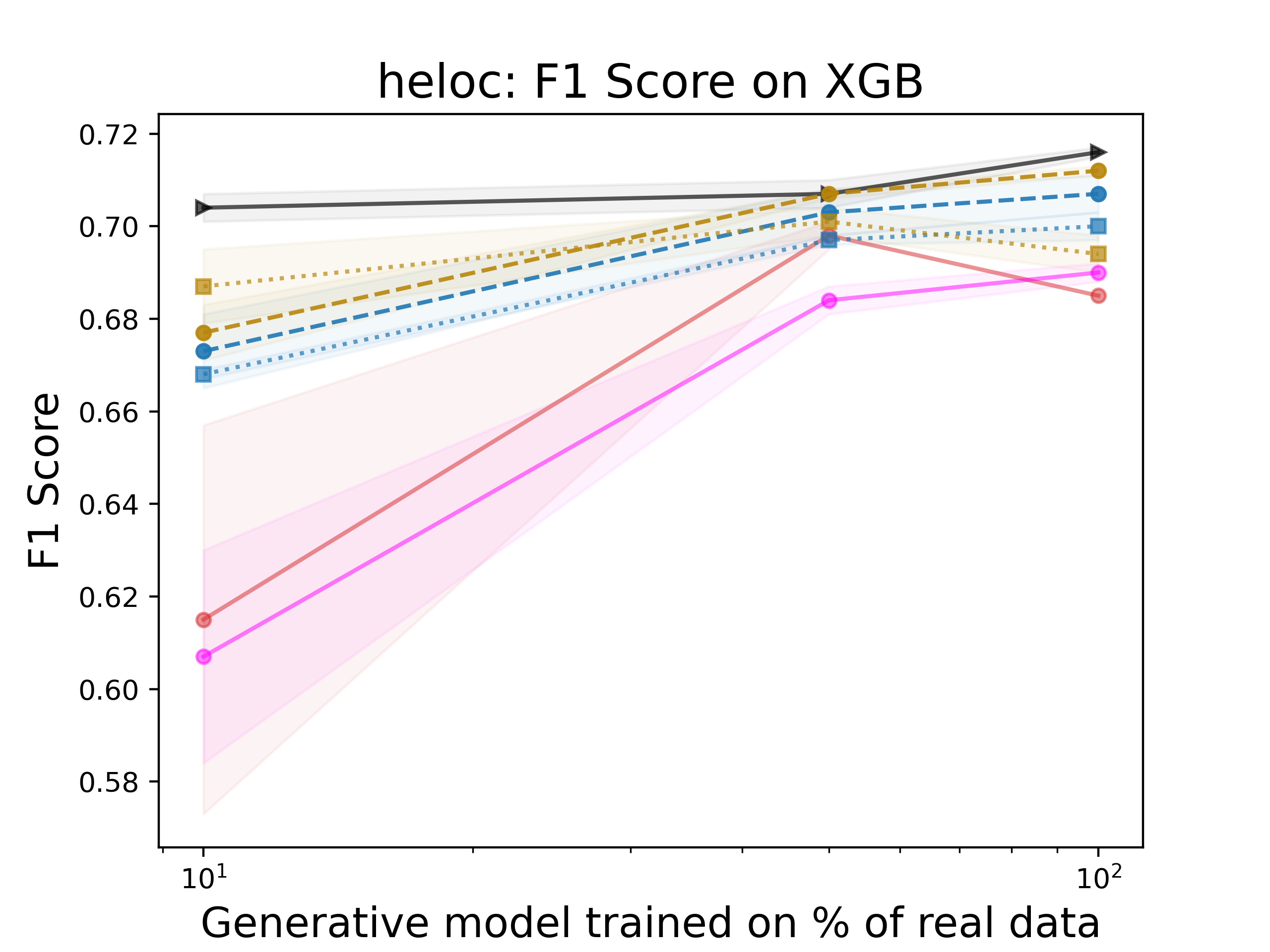}
    \end{minipage}
     \begin{minipage}[t]{0.245\textwidth}
        \centering
        \includegraphics[width=\linewidth]{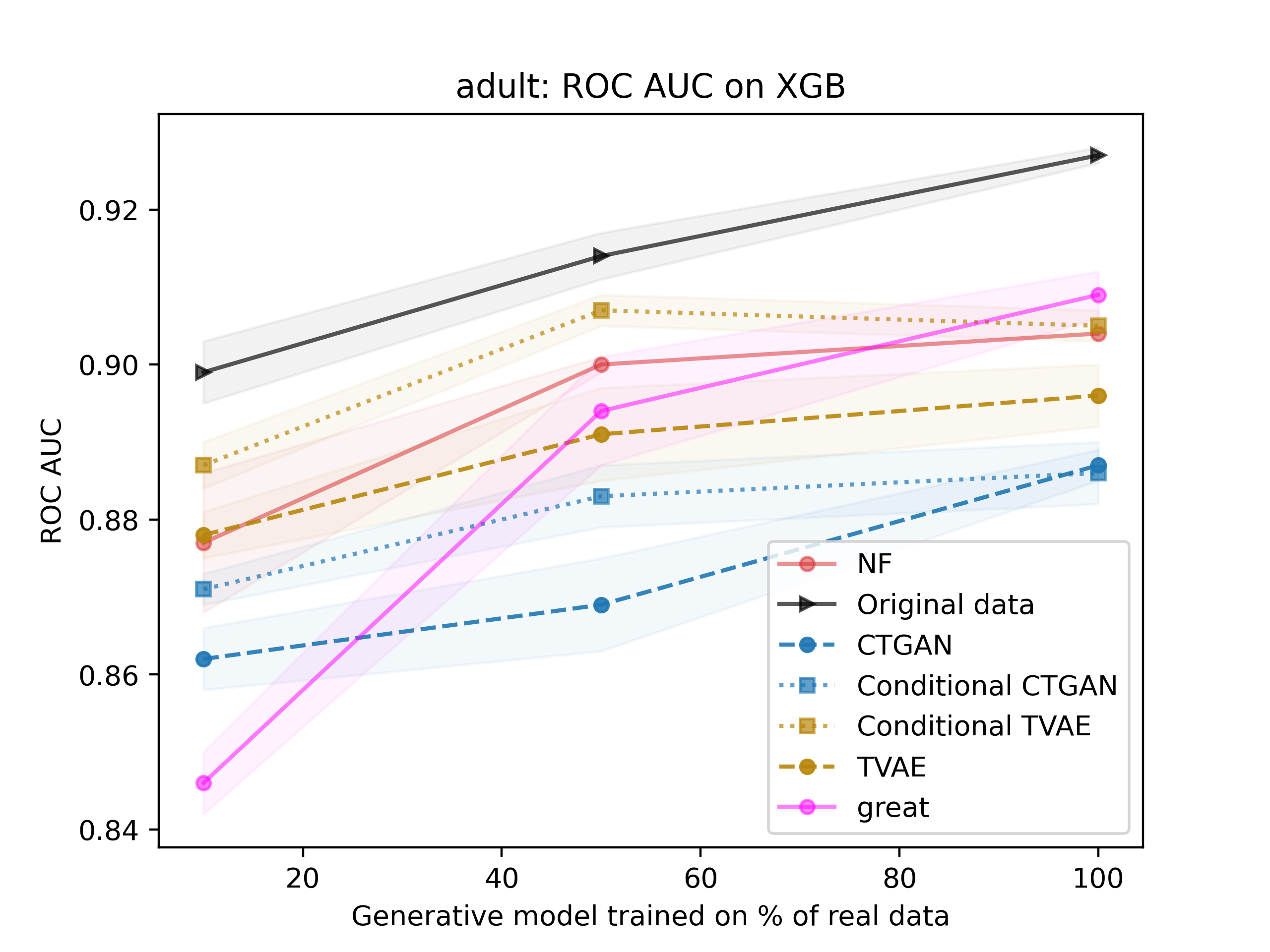}
    \end{minipage}%
    \hfill
    \begin{minipage}[t]{0.245\textwidth}
        \centering
        \includegraphics[width=\linewidth]{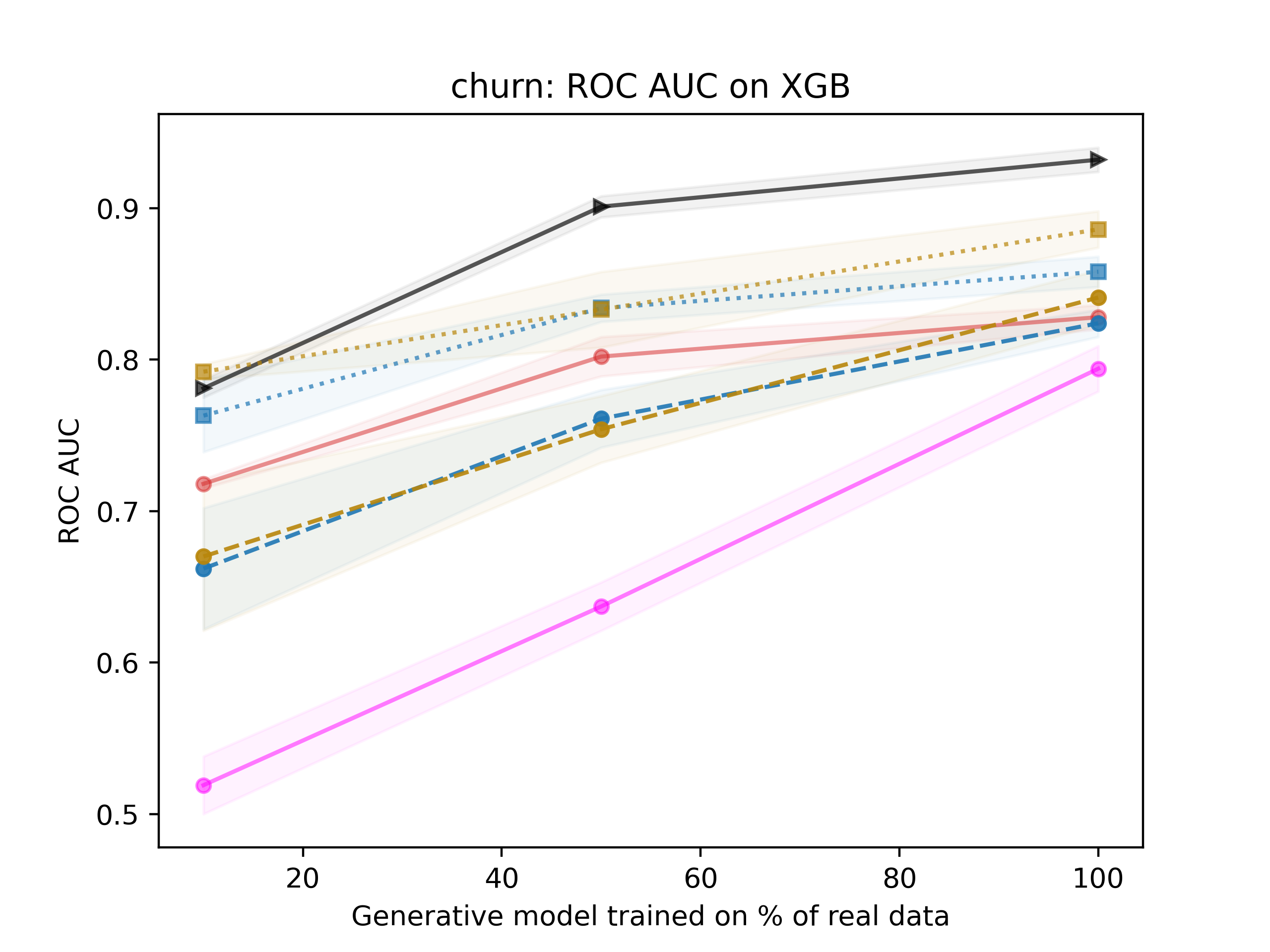}
    \end{minipage}
    \hfill
    \begin{minipage}[t]{0.245\textwidth}
        \centering
        \includegraphics[width=\linewidth]{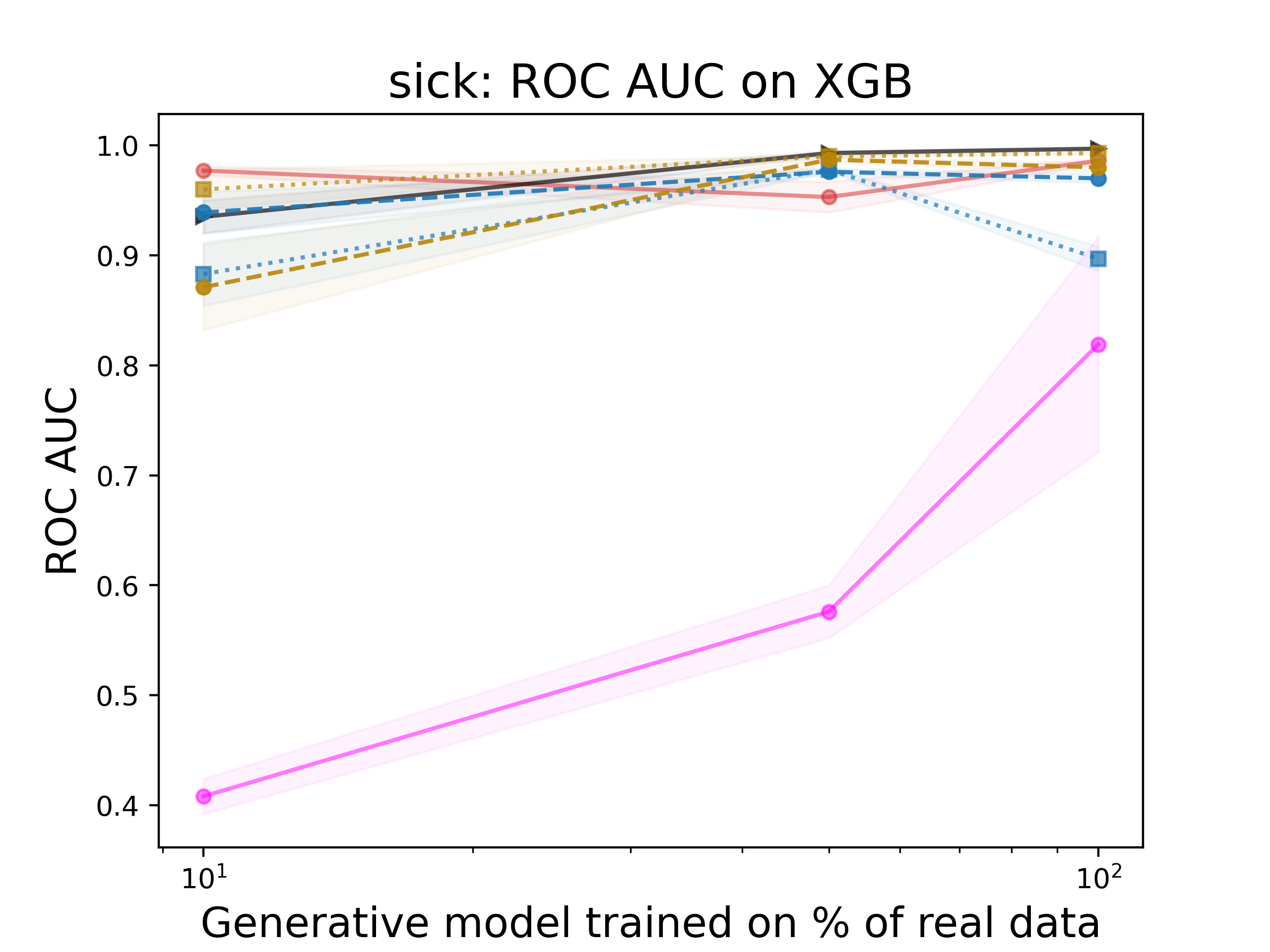}
    \end{minipage}
    \hfill
    \begin{minipage}[t]{0.245\textwidth}
        \centering
        \includegraphics[width=\linewidth]{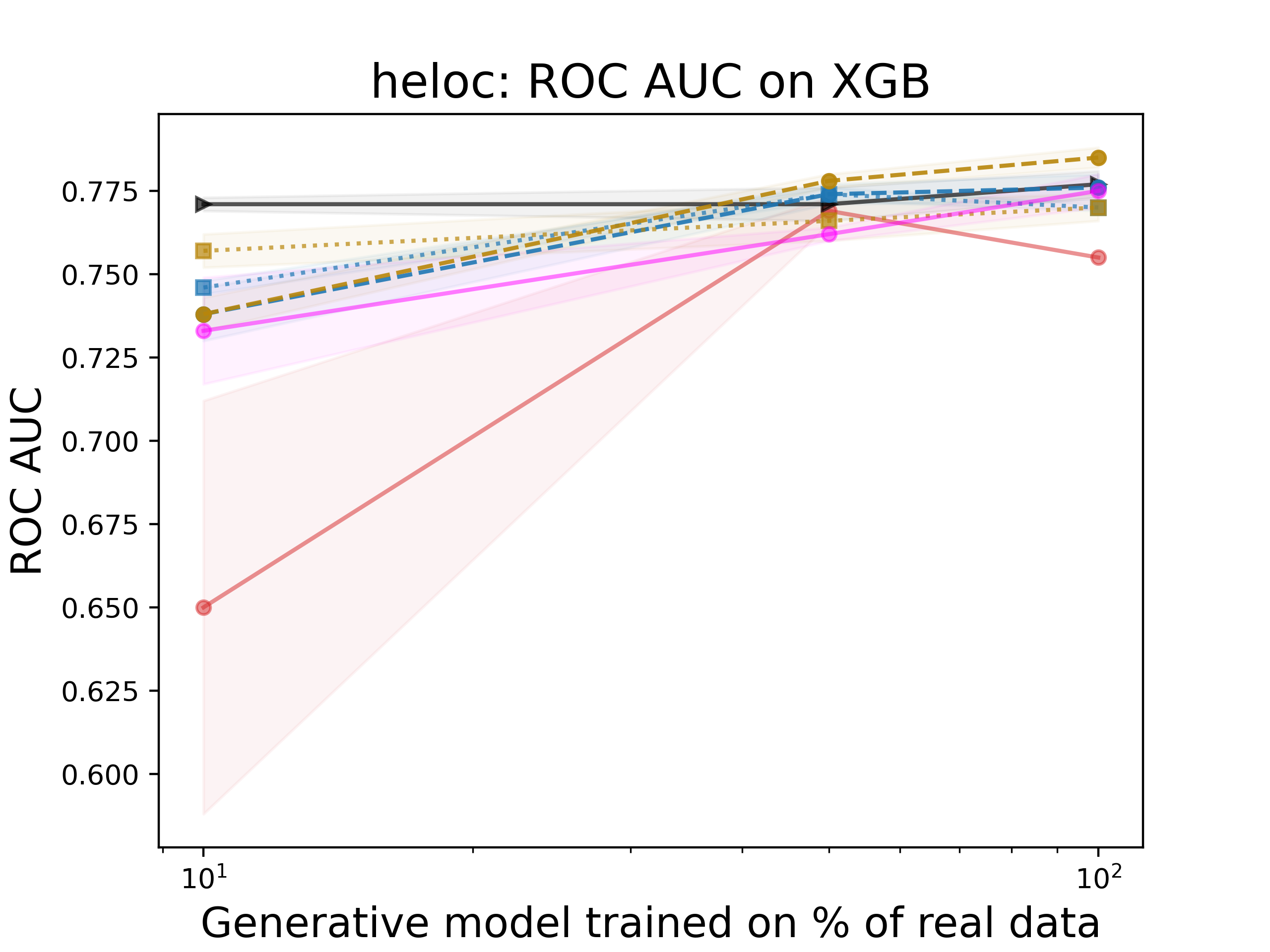}
    \end{minipage}
    \caption{(\emph{Synthetic data quality experiment}) F1-score and ROC AUC on a downstream XGB model for all classification datasets.}
    \label{fig:appendix-synthetic-classification}
\end{figure*}

\begin{figure*}[!t]
        \centering
        \begin{minipage}[t]{0.325\textwidth}
        \includegraphics[width=\linewidth]{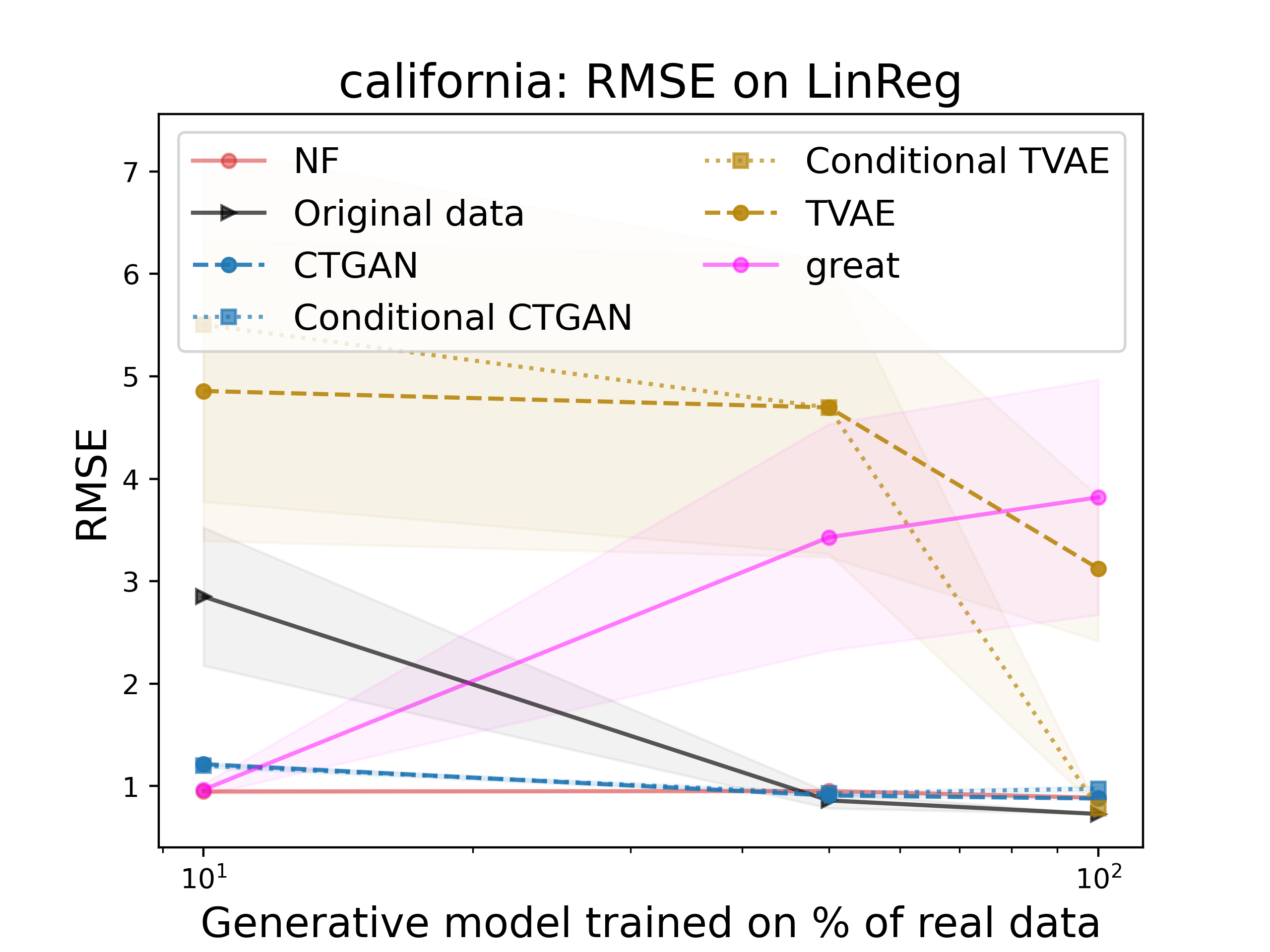}
    \end{minipage}%
    \begin{minipage}[t]{0.325\textwidth}
        \includegraphics[width=\linewidth]{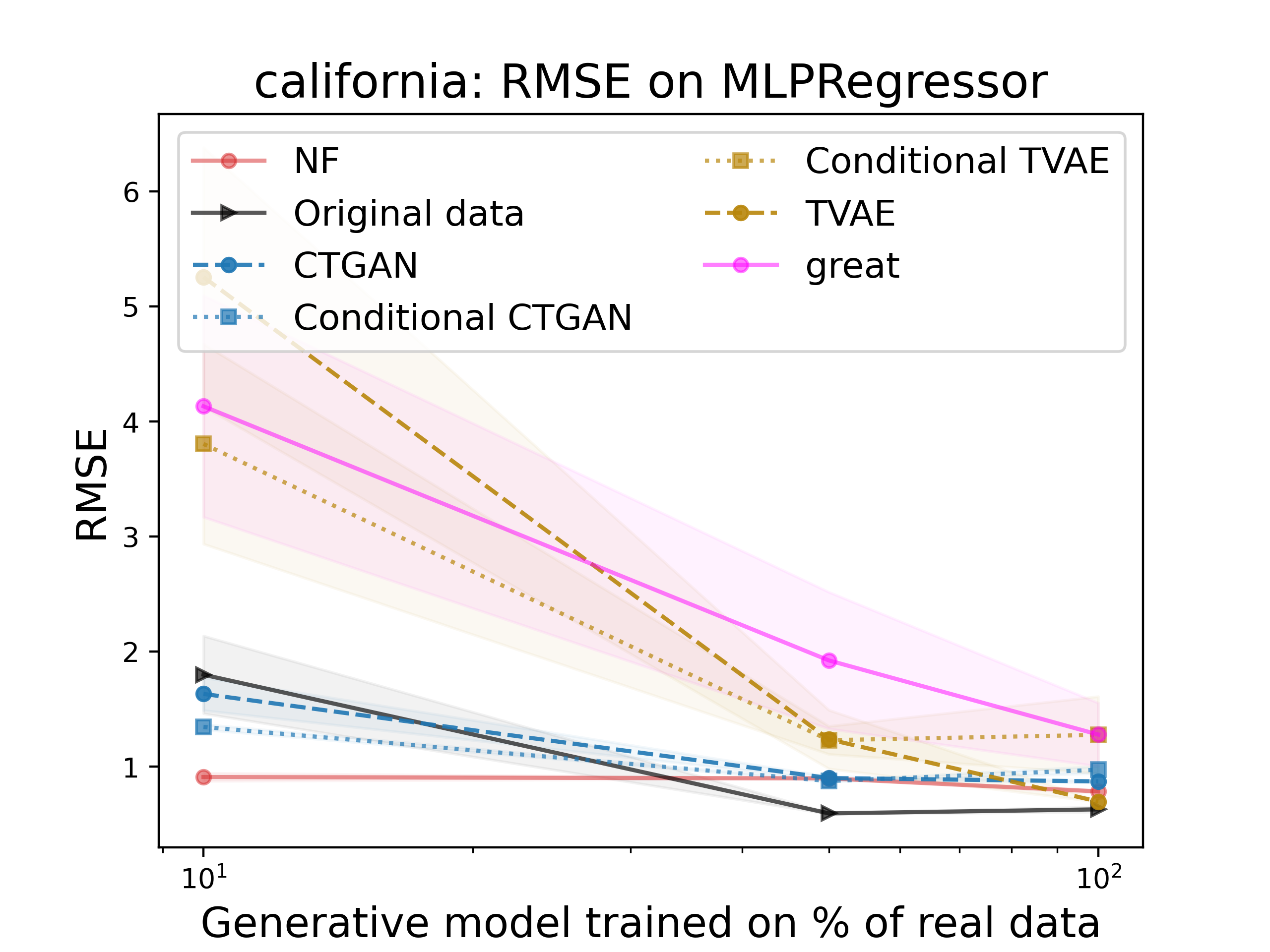}
    \end{minipage}
    \caption{(\emph{Synthetic data quality}) RMSE for downstream regression tasks on a linear model and a multi-layered perceptron.}
        \label{fig:appendix-synthetic-regression}
\end{figure*}

\begin{figure*}[!t]
        \begin{minipage}[t]{0.245\textwidth}
        \centering
        \includegraphics[width=\linewidth]{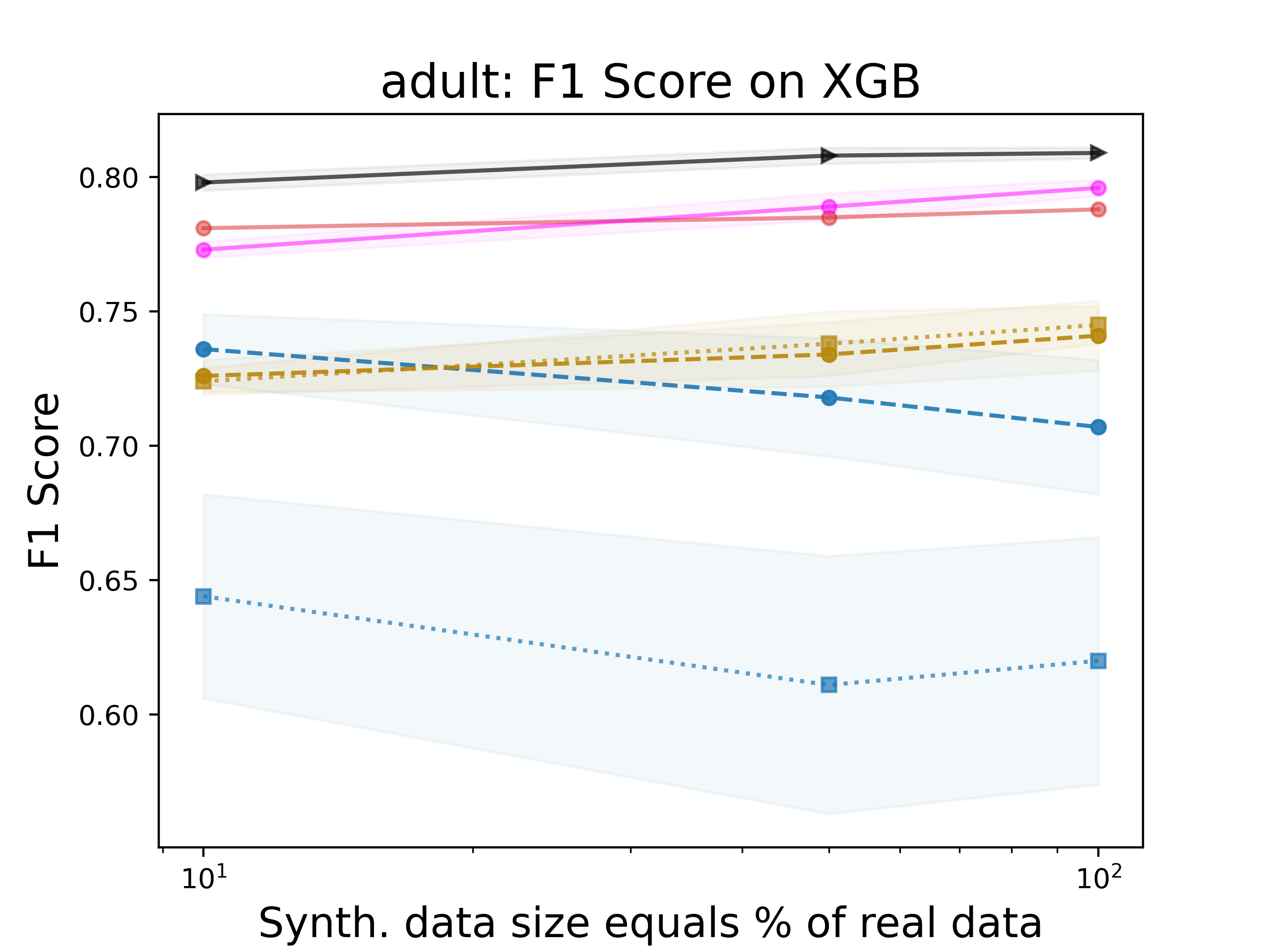}
    \end{minipage}%
    \hfill
    \begin{minipage}[t]{0.245\textwidth}
        \centering
        \includegraphics[width=\linewidth]{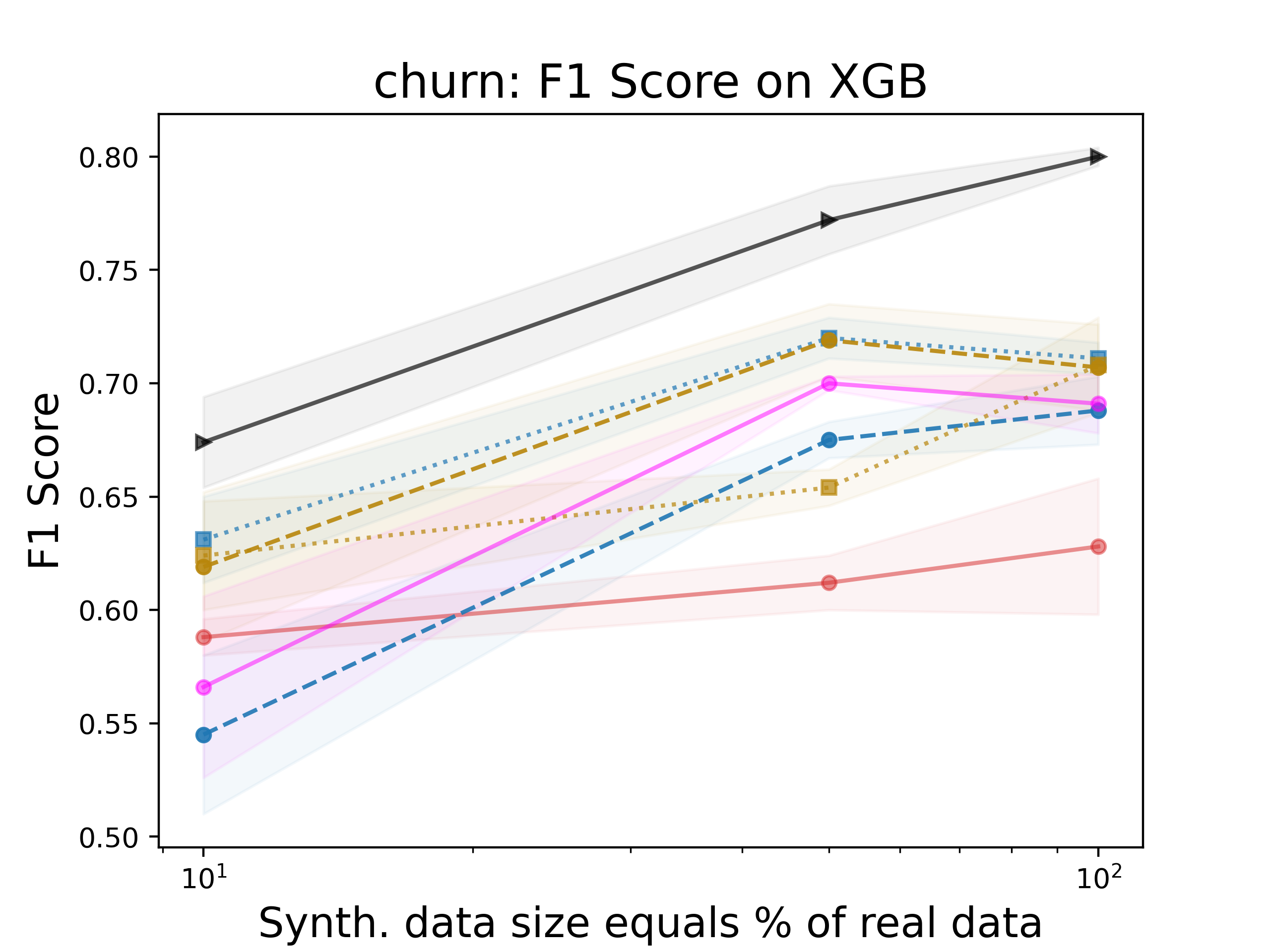}
    \end{minipage}
    \hfill
    \begin{minipage}[t]{0.245\textwidth}
        \centering
        \includegraphics[width=\linewidth]{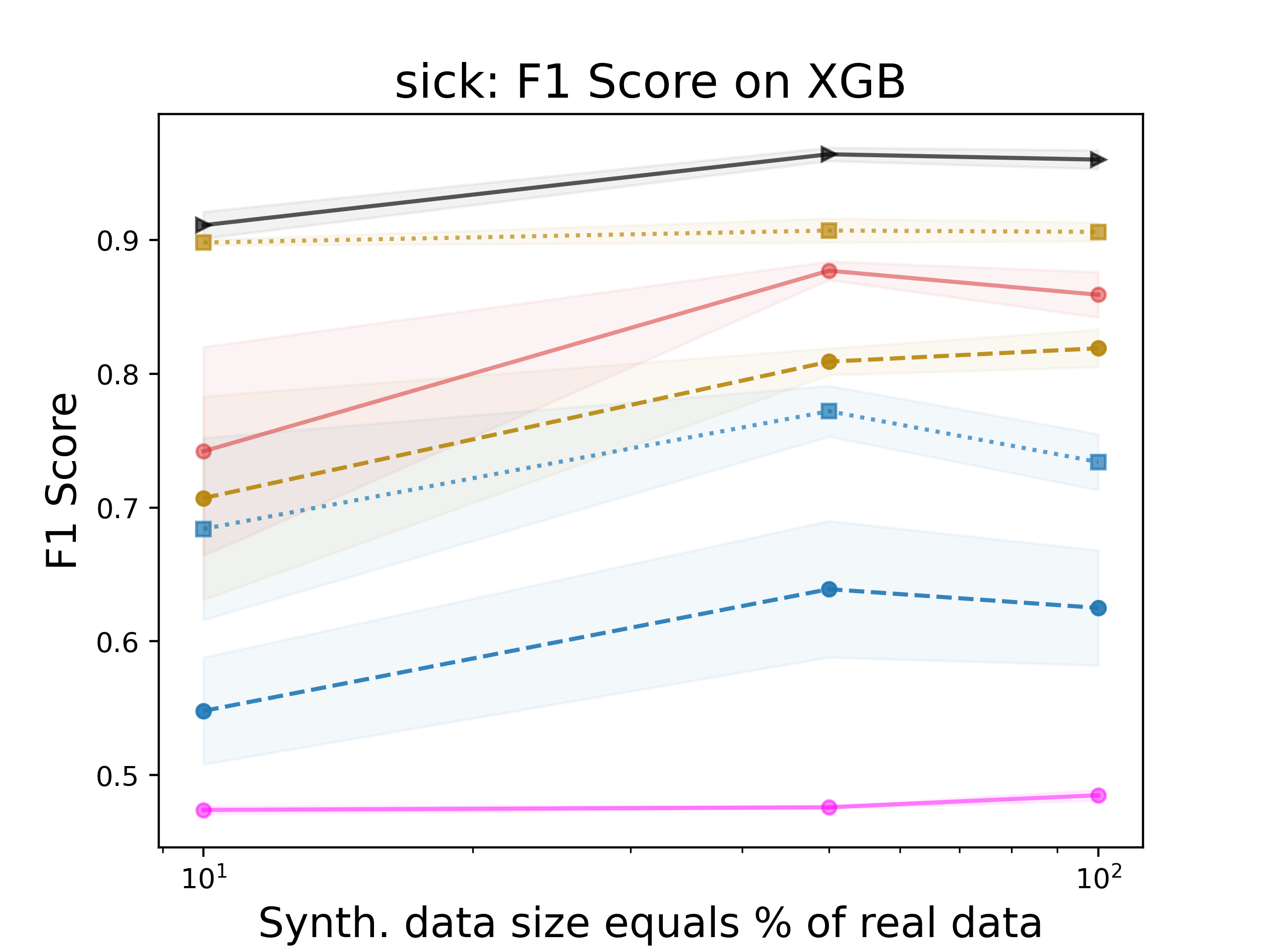}
    \end{minipage}
    \hfill
    \begin{minipage}[t]{0.245\textwidth}
        \centering
        \includegraphics[width=\linewidth]{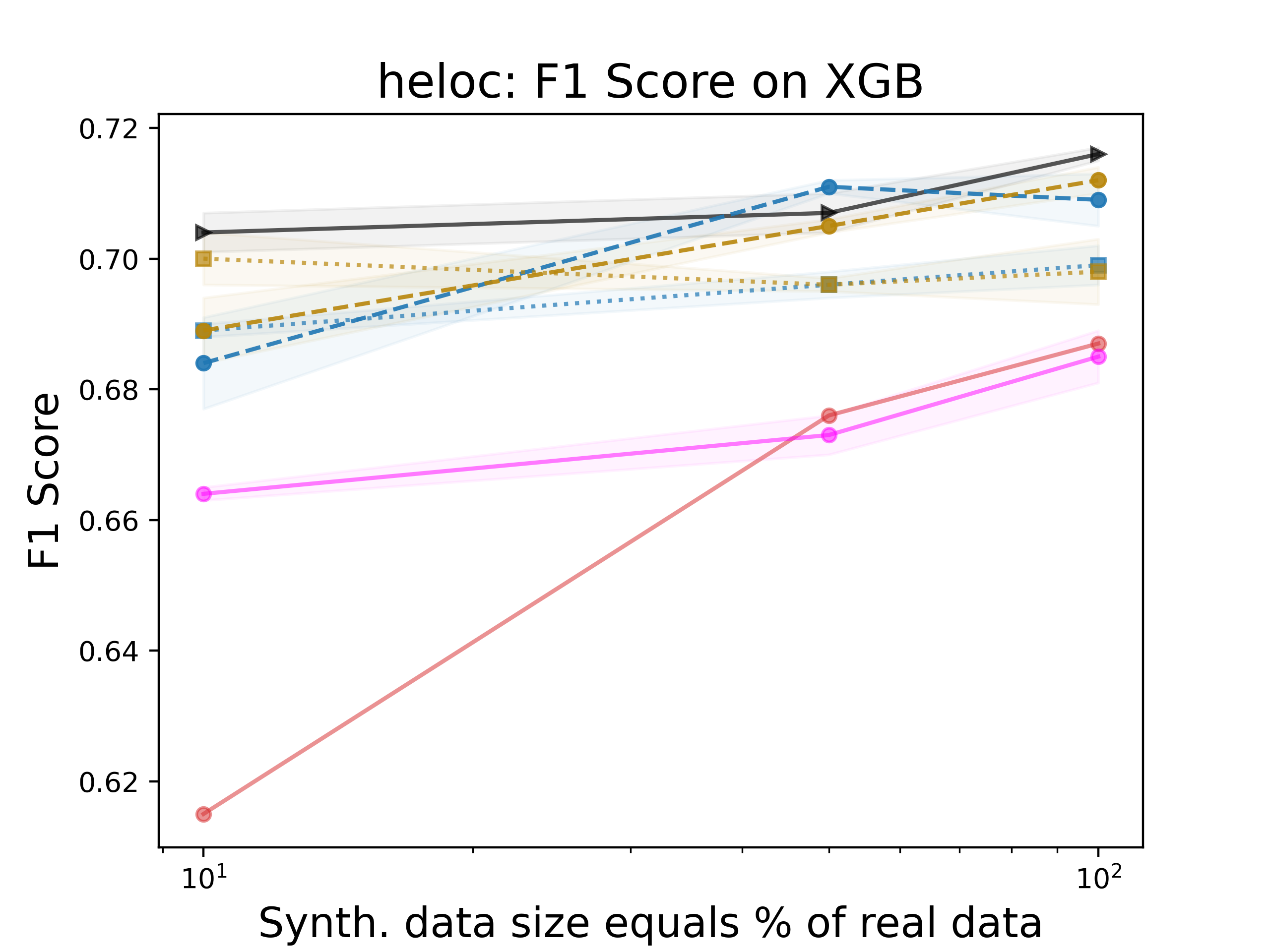}
    \end{minipage}
        \begin{minipage}[t]{0.245\textwidth}
        \centering
        \includegraphics[width=\linewidth]{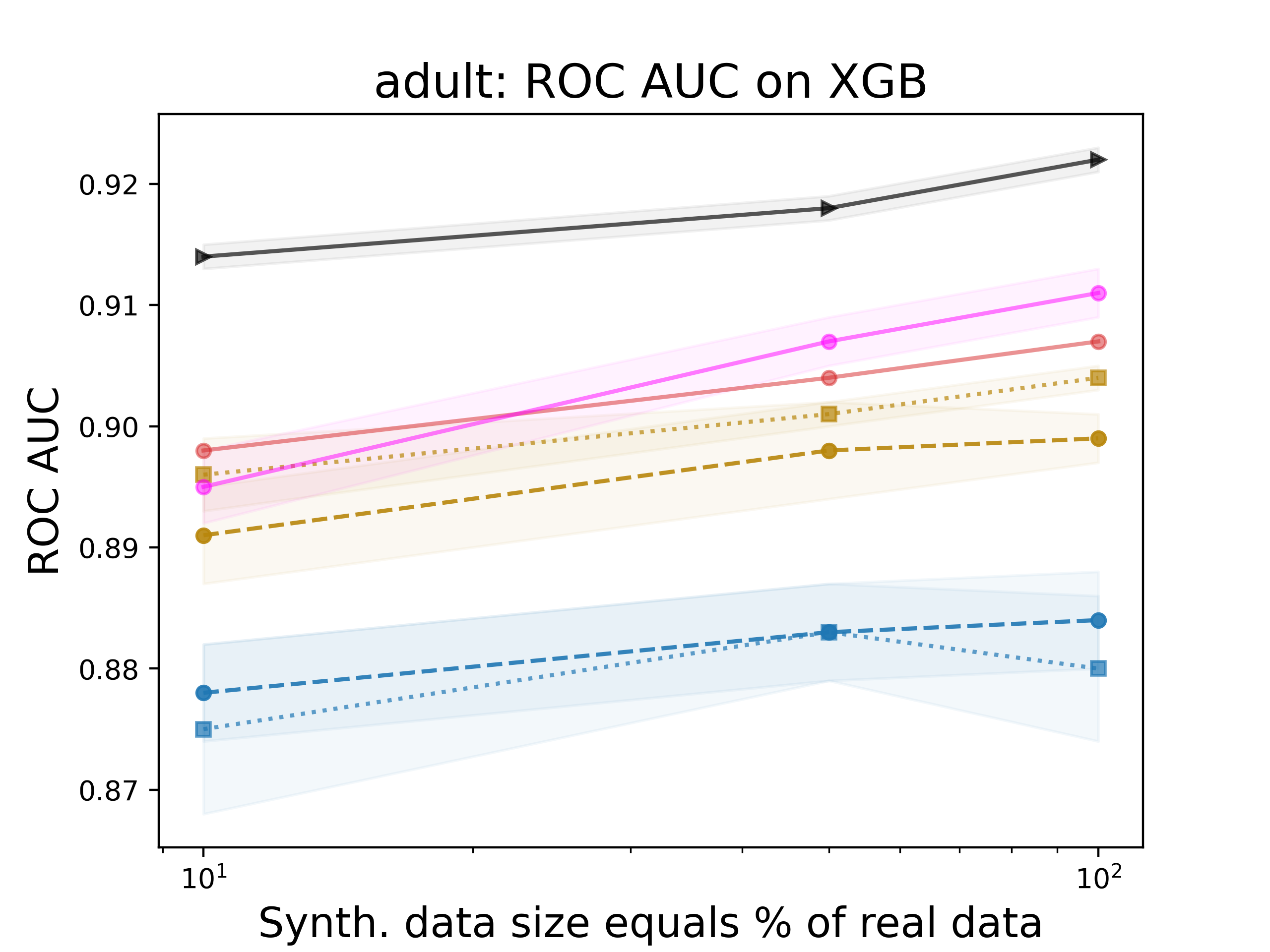}
    \end{minipage}%
    \hfill
    \begin{minipage}[t]{0.245\textwidth}
        \centering
        \includegraphics[width=\linewidth]{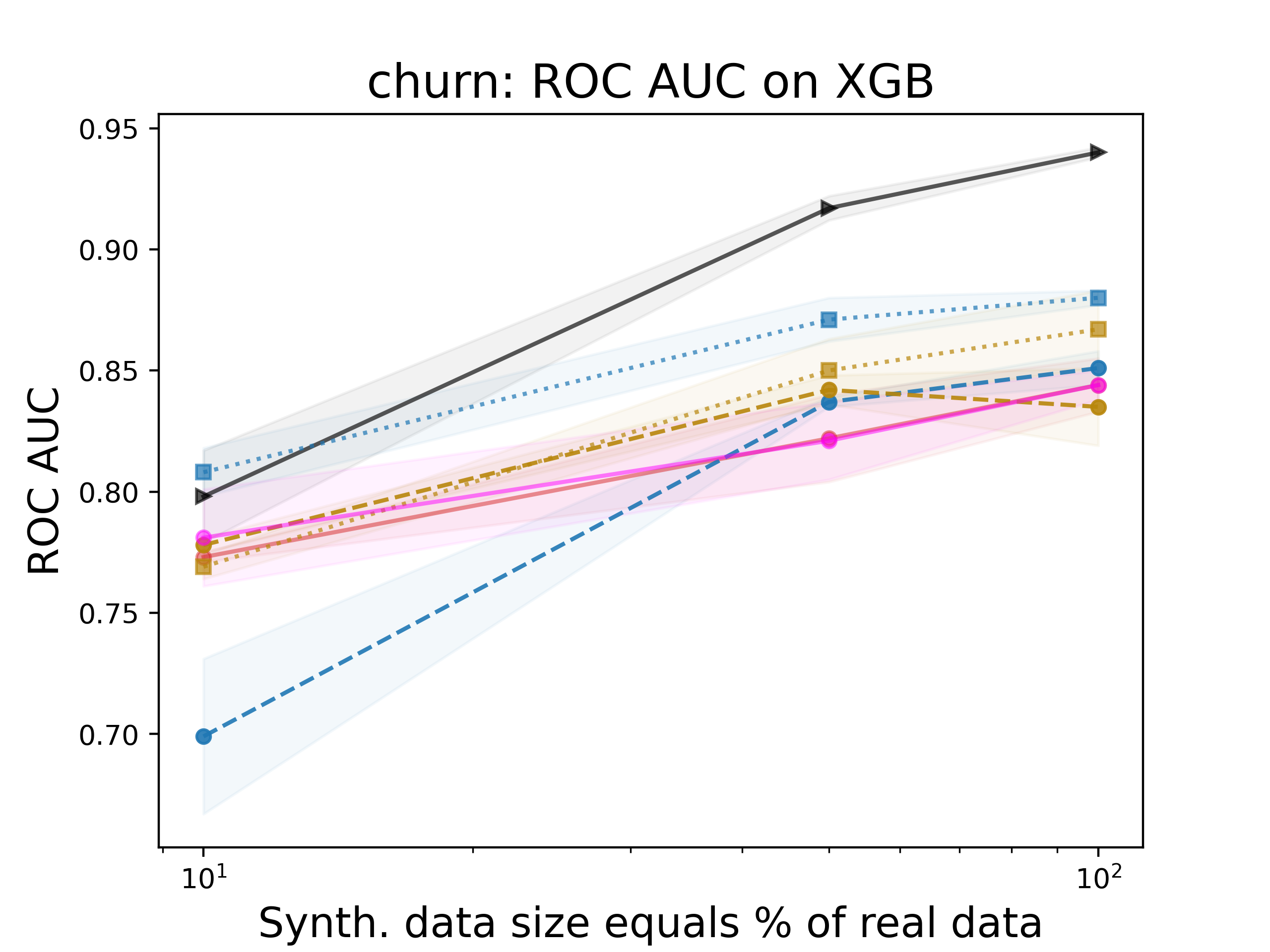}
    \end{minipage}
    \hfill
    \begin{minipage}[t]{0.245\textwidth}
        \centering
        \includegraphics[width=\linewidth]{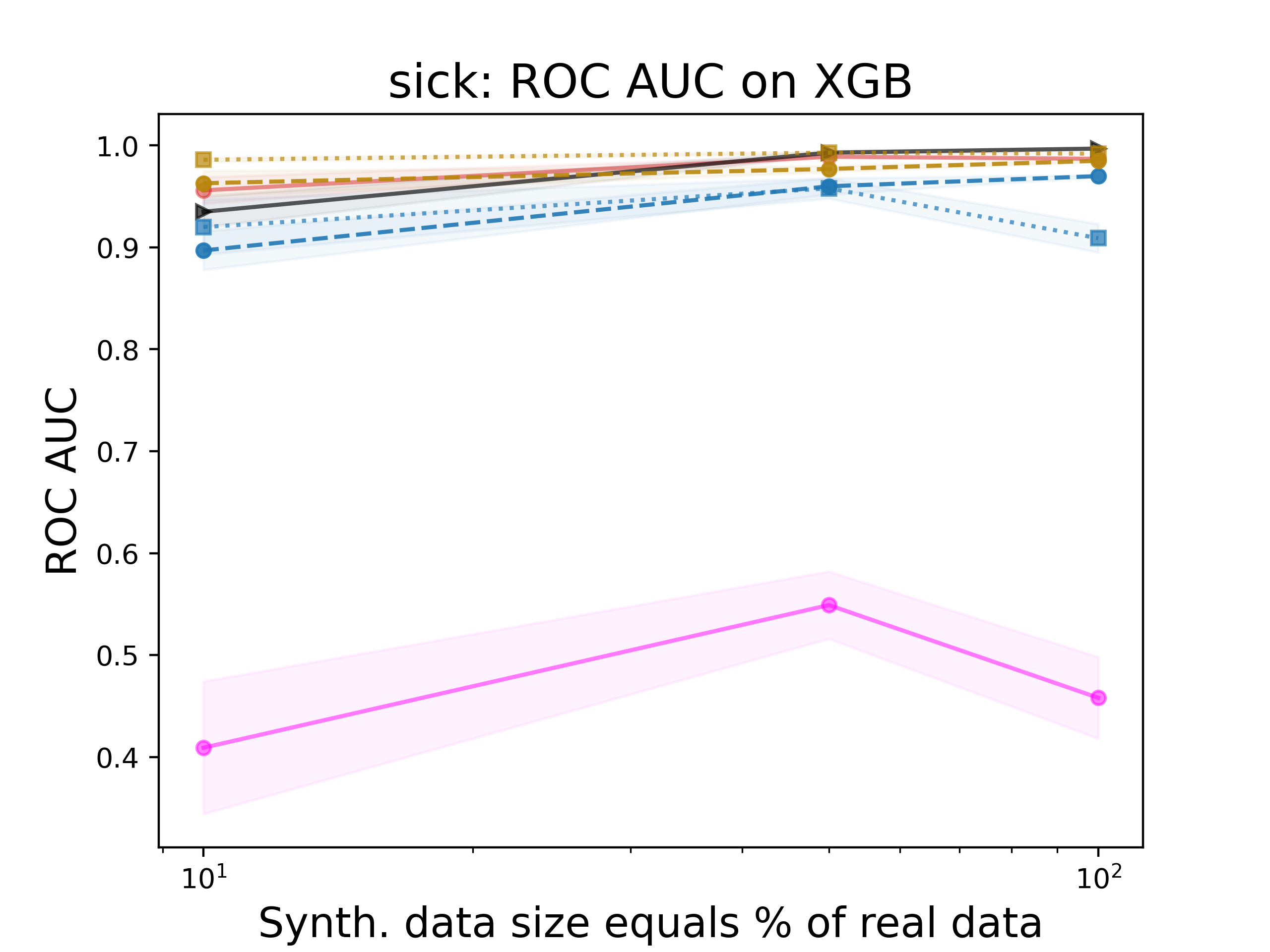}
    \end{minipage}
    \hfill
    \begin{minipage}[t]{0.245\textwidth}
        \centering
        \includegraphics[width=\linewidth]{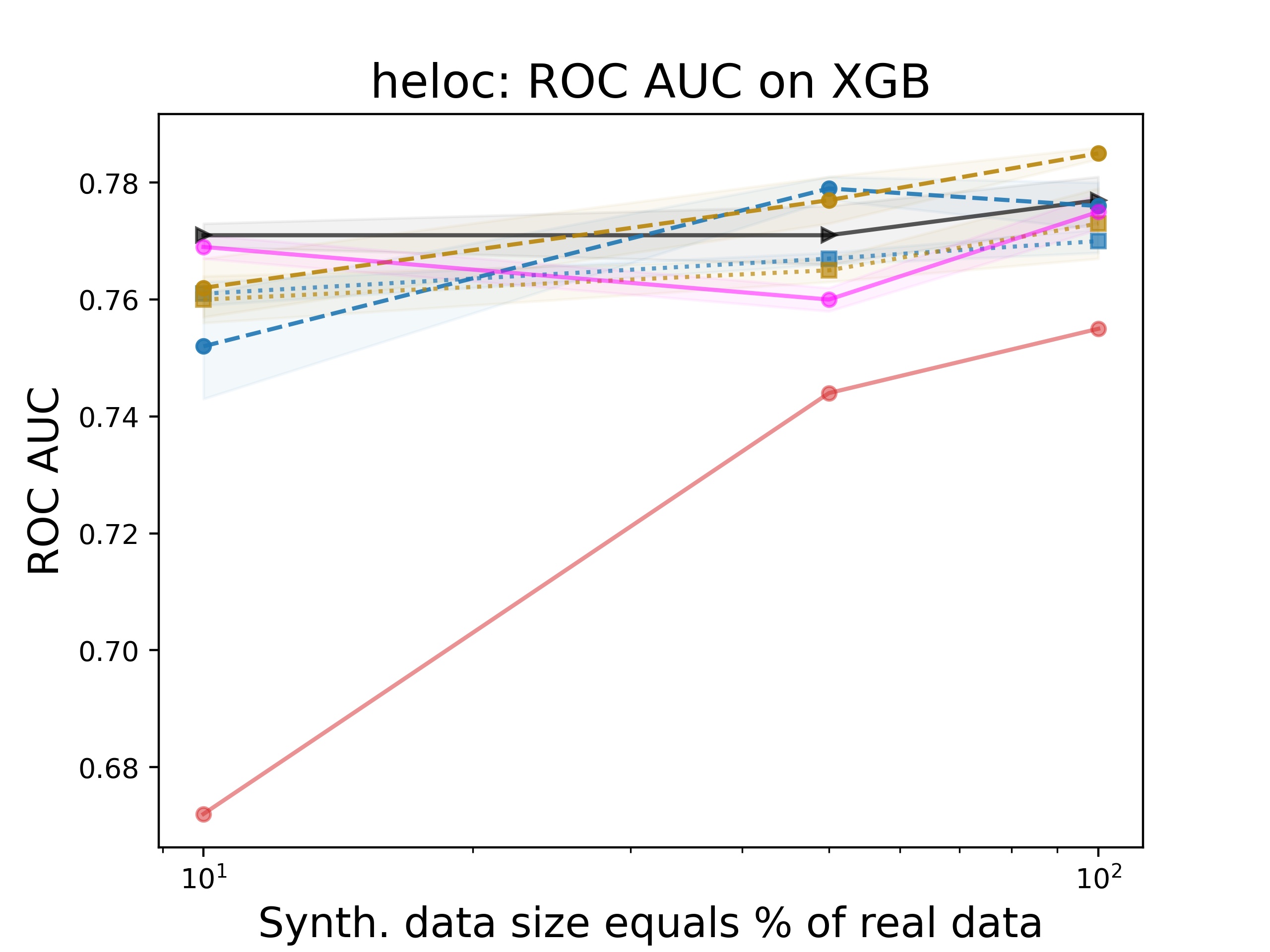}
    \end{minipage}
    \caption{(\emph{Data summarization experiment}) F1-score and ROC AUC on a downstream XGB model for all classification datasets.}
    \label{fig:appendix-class-summarization}
\end{figure*}

\begin{figure*}[!t]
        \centering
        \begin{minipage}[t]{0.325\textwidth}
        \includegraphics[width=\linewidth]{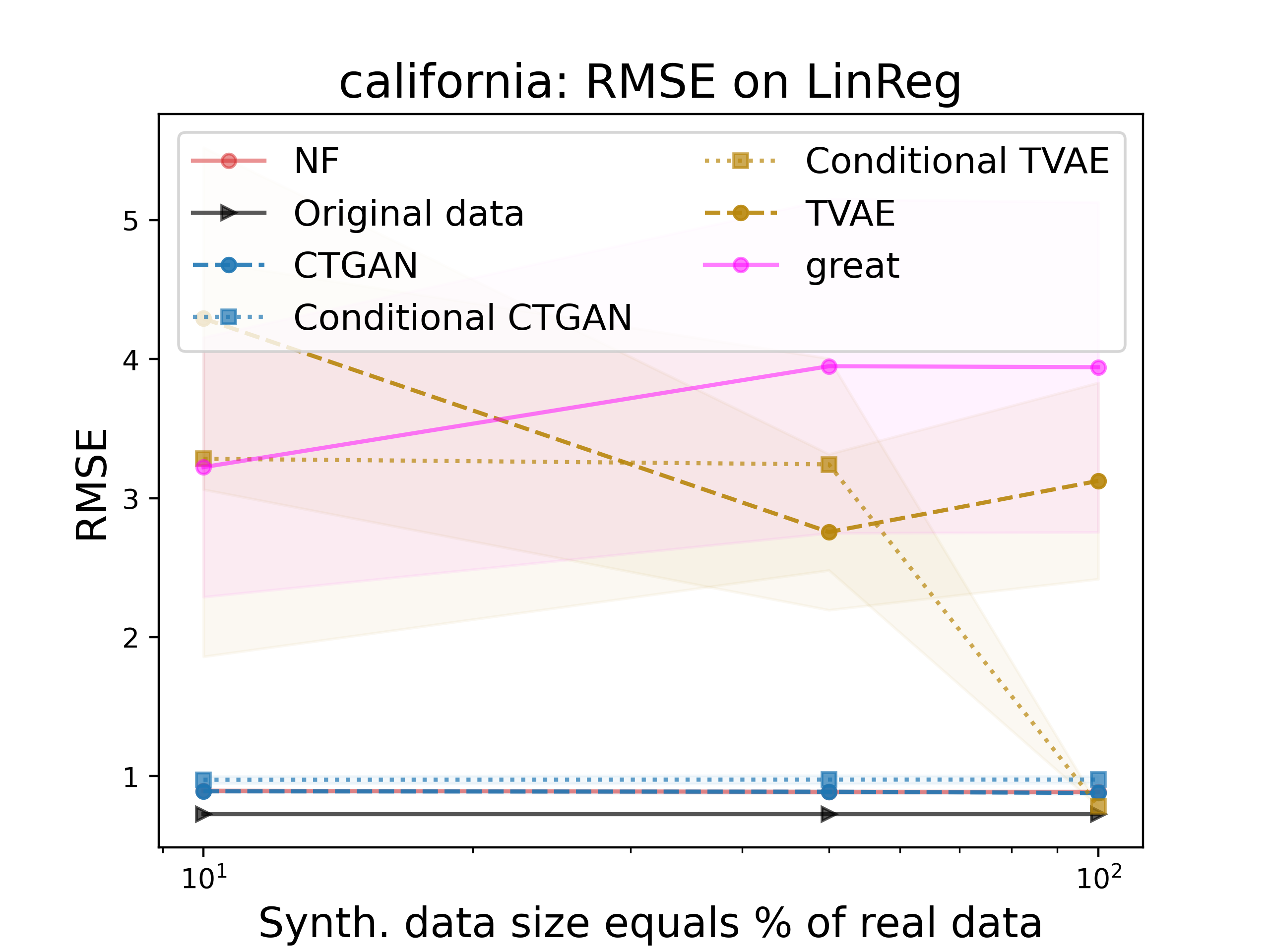}
        \end{minipage}%
        \begin{minipage}[t]{0.325\textwidth}
        \includegraphics[width=\linewidth]{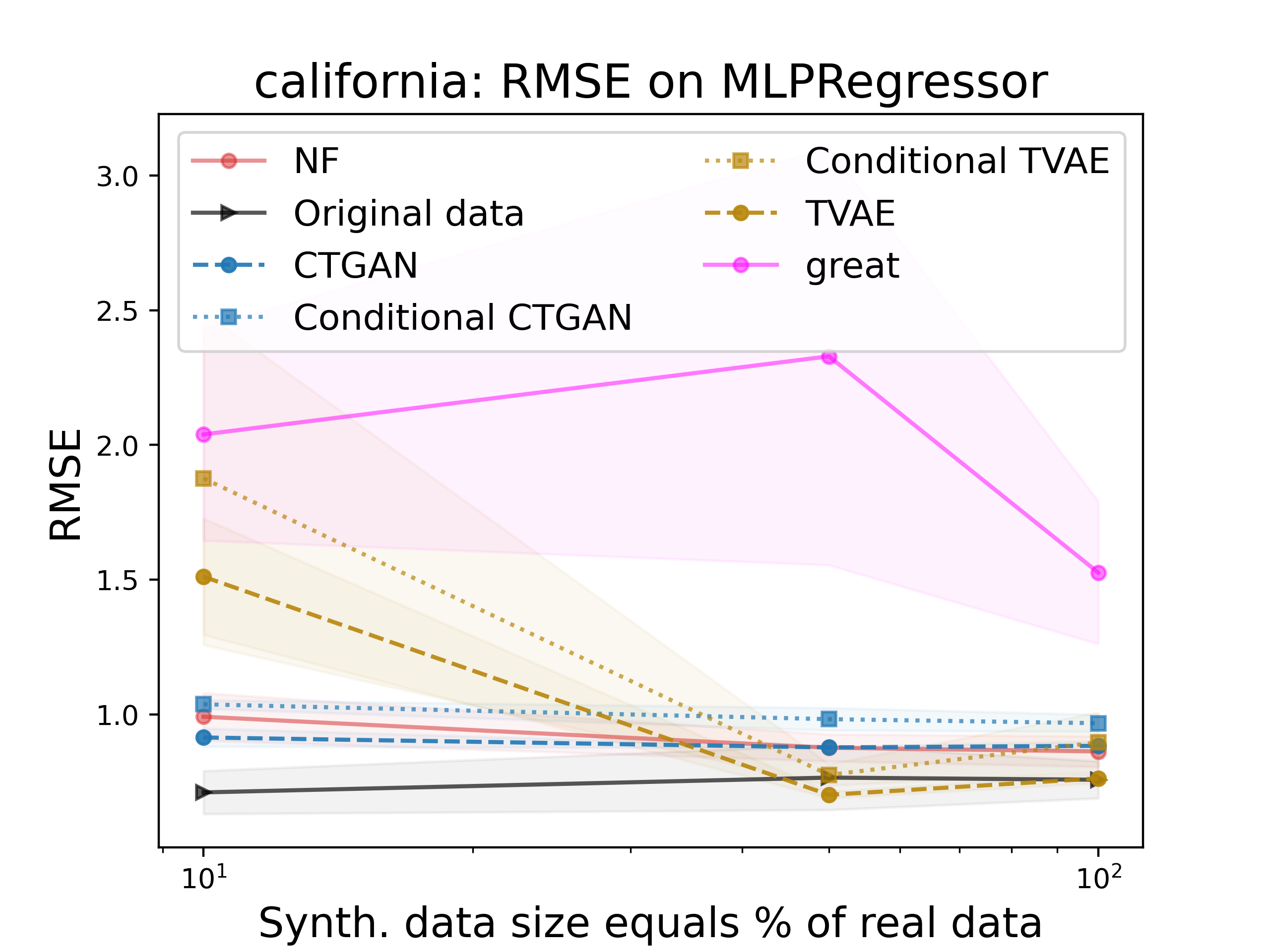}
    \end{minipage}
    \caption{(\emph{Data summarization experiment}) Predictive metrics for downstream regression tasks on a linear model and a multi-layered perceptron.}
        \label{fig:appendix-summarization-regression}
\end{figure*}

\begin{figure*}[!t]
        \begin{minipage}[t]{0.245\textwidth}
        \centering
        \includegraphics[width=\linewidth]{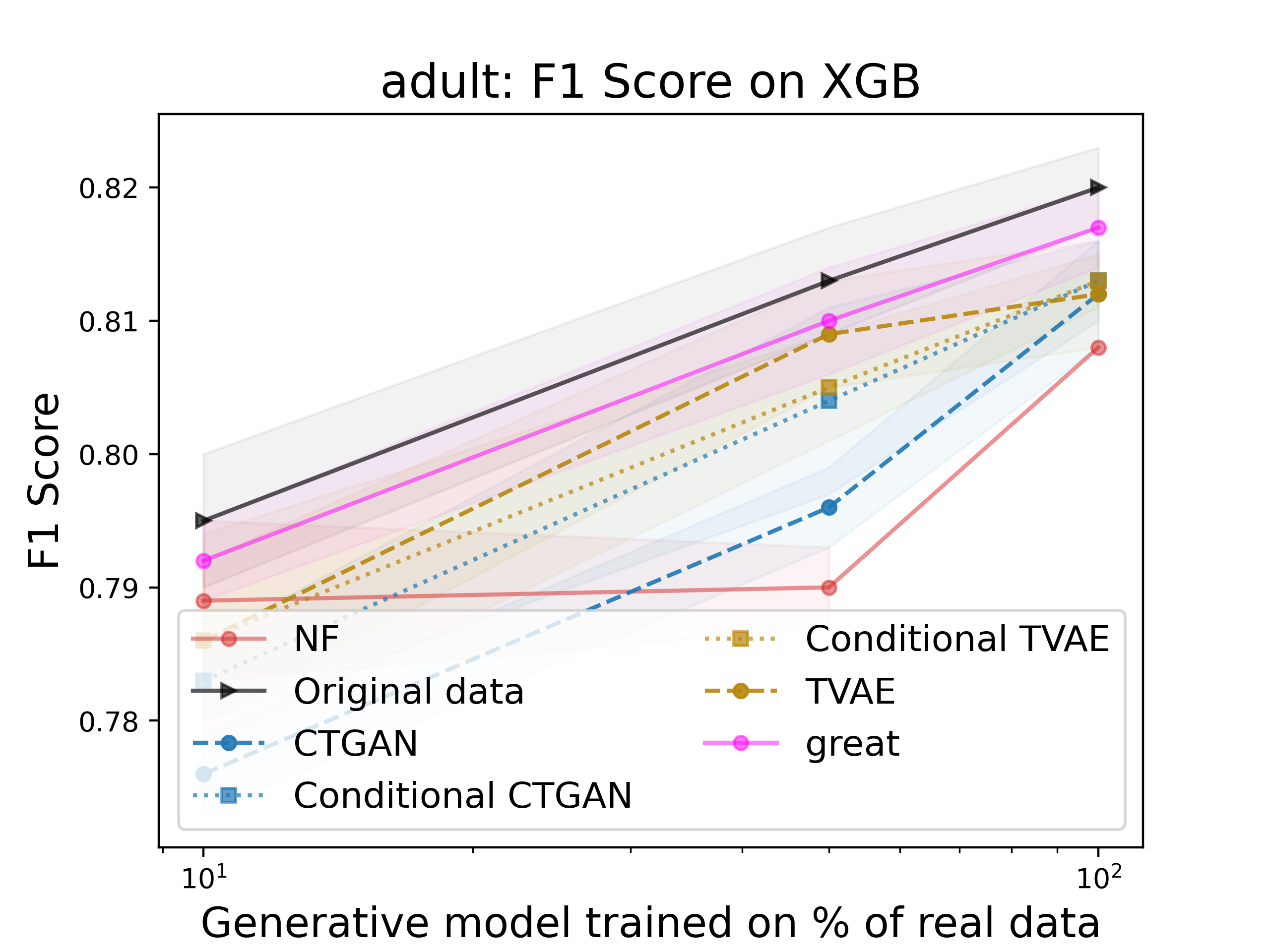}
    \end{minipage}%
    \hfill
    \begin{minipage}[t]{0.245\textwidth}
        \centering
        \includegraphics[width=\linewidth]{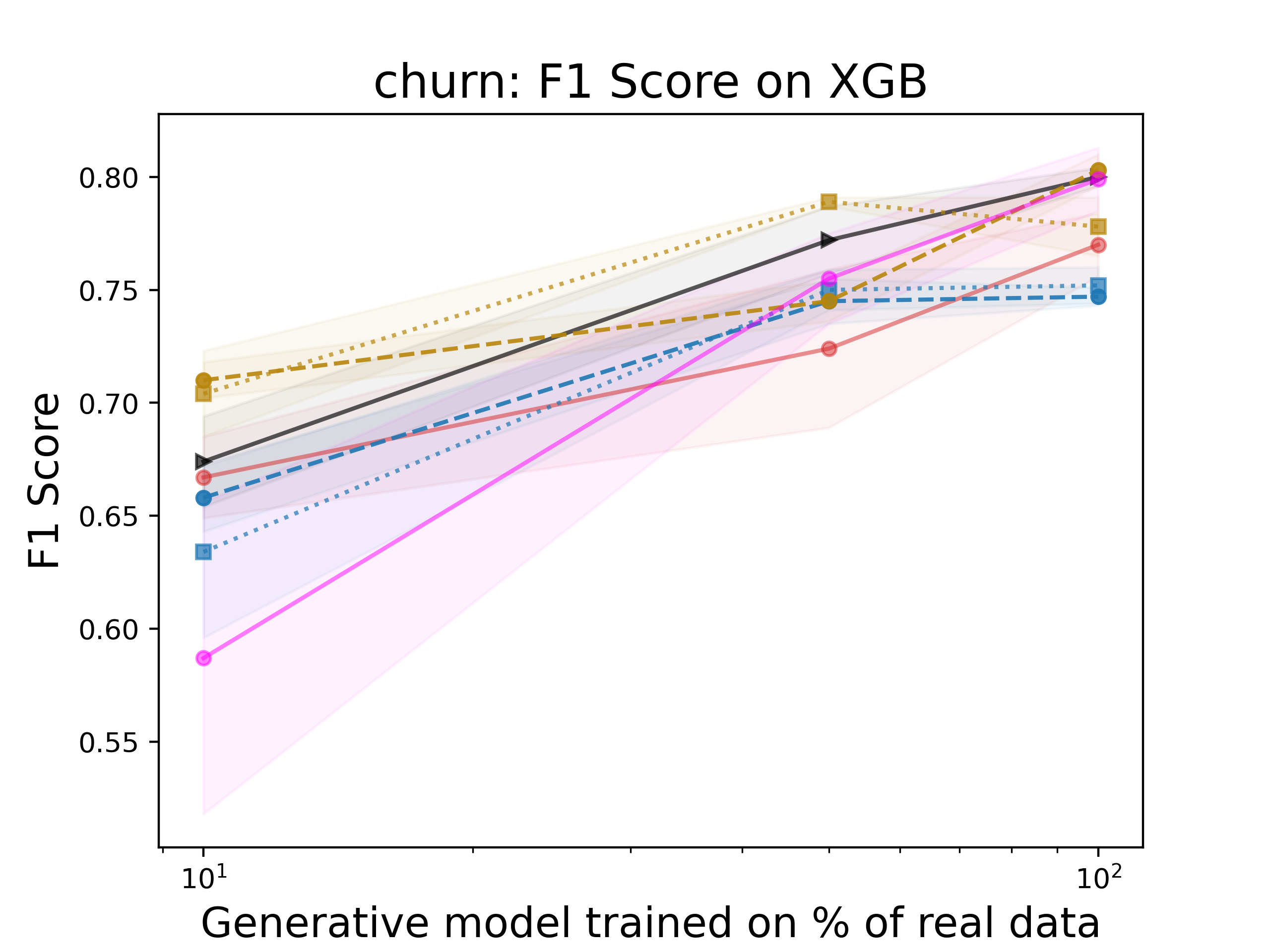}
    \end{minipage}
    \hfill
    \begin{minipage}[t]{0.245\textwidth}
        \centering
        \includegraphics[width=\linewidth]{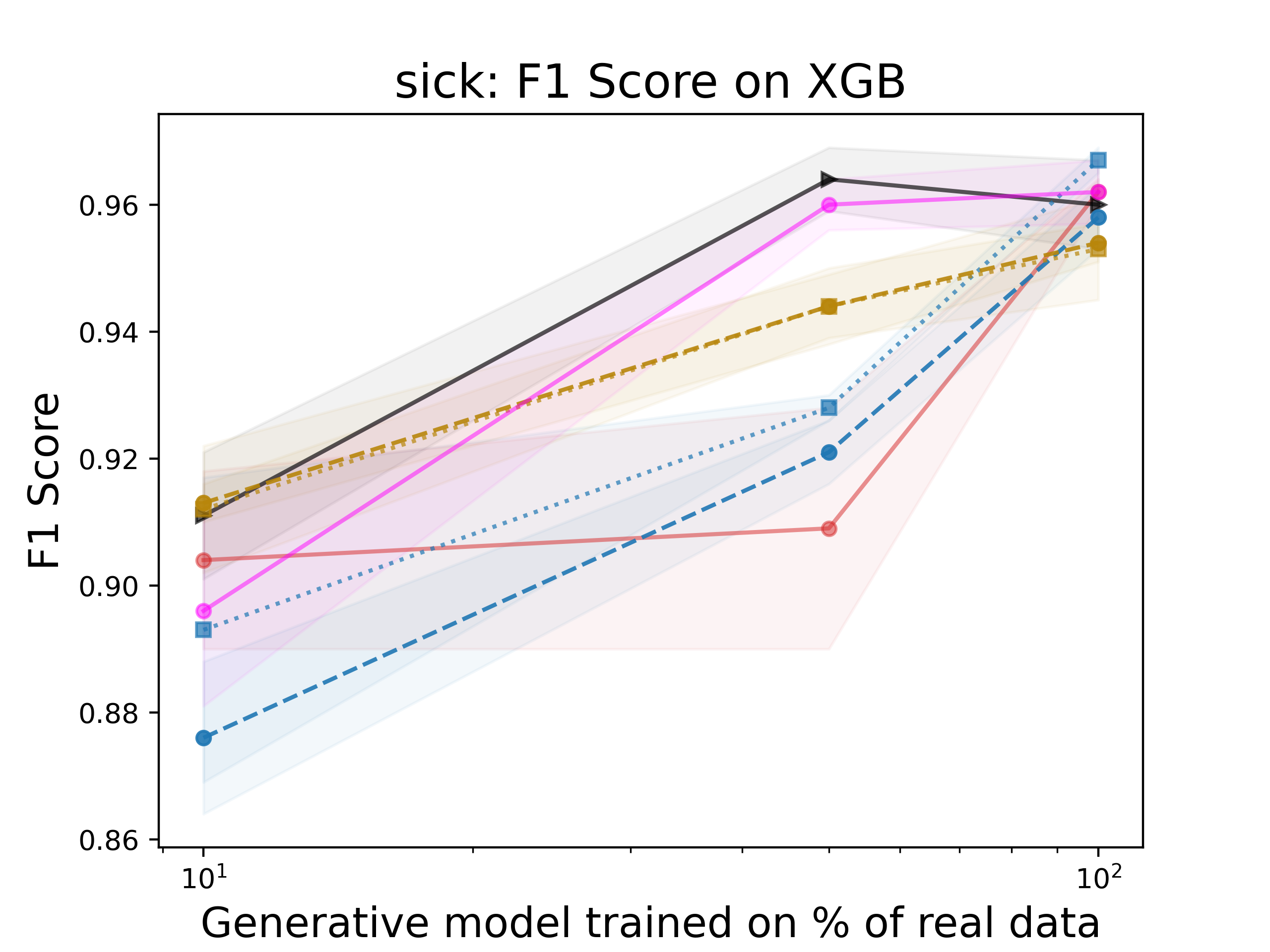}
    \end{minipage}
    \hfill
    \begin{minipage}[t]{0.245\textwidth}
        \centering
        \includegraphics[width=\linewidth]{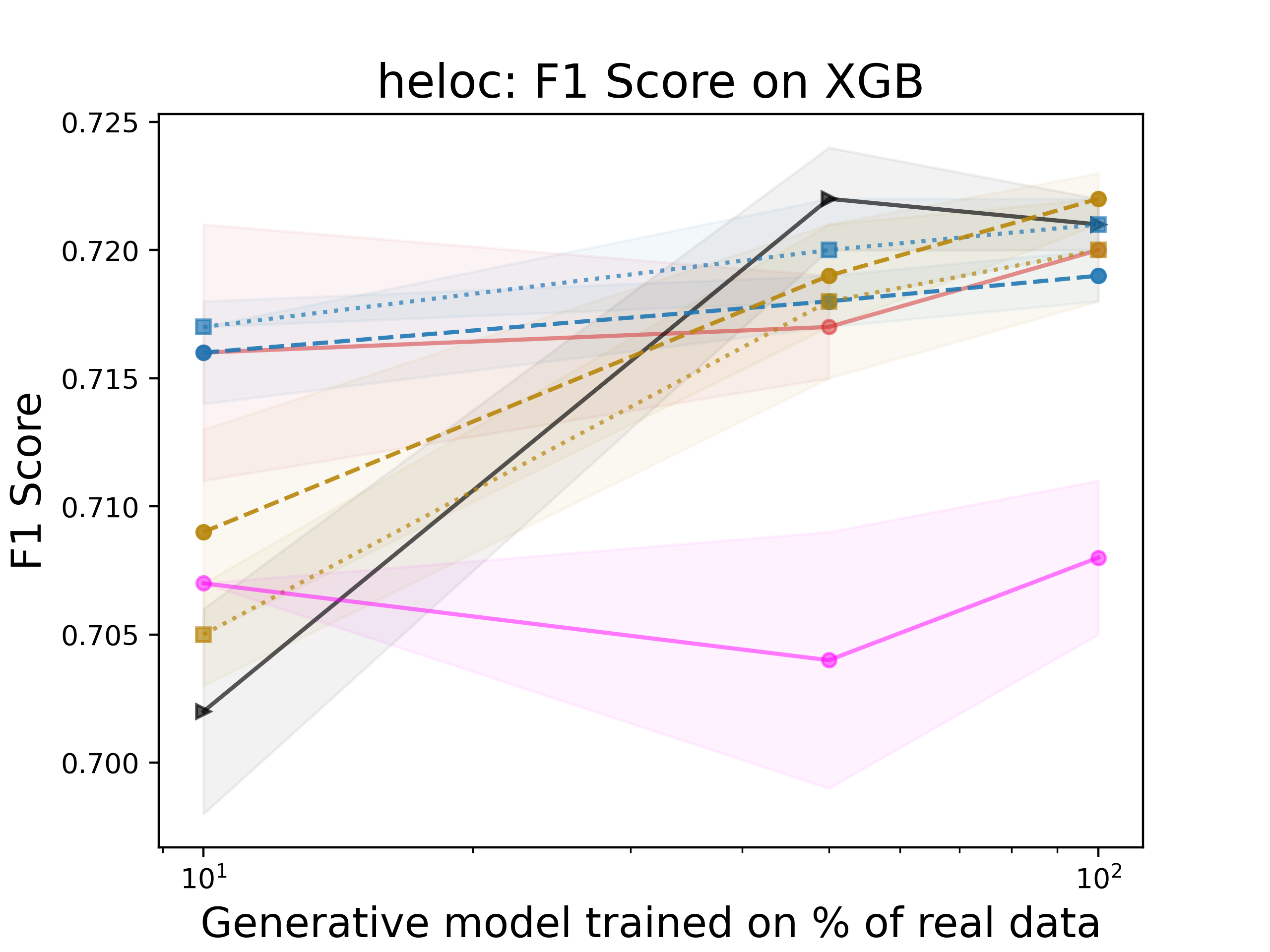}
    \end{minipage}
        \begin{minipage}[t]{0.245\textwidth}
        \centering
        \includegraphics[width=\linewidth]{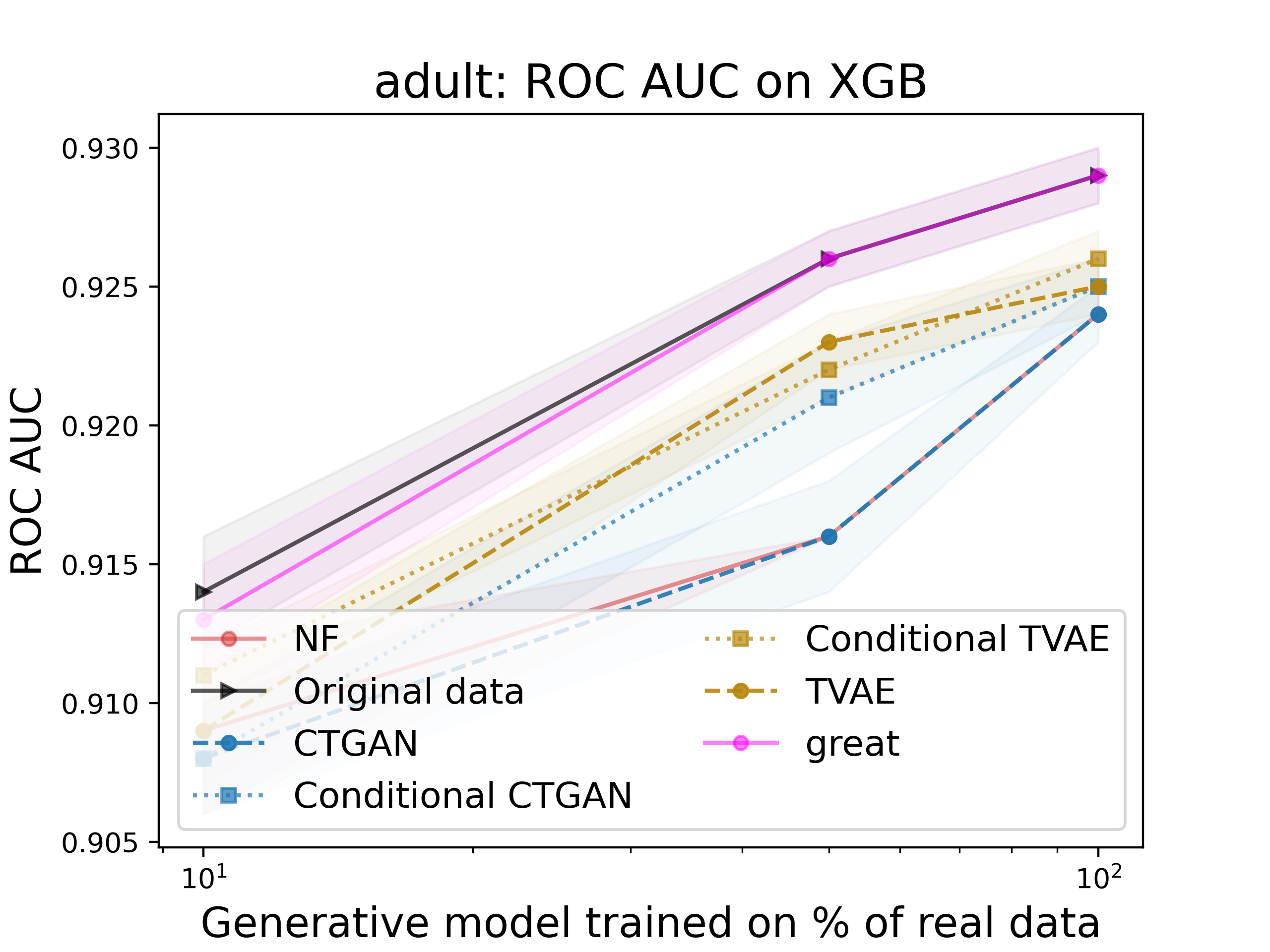}
    \end{minipage}%
    \hfill
    \begin{minipage}[t]{0.245\textwidth}
        \centering
        \includegraphics[width=\linewidth]{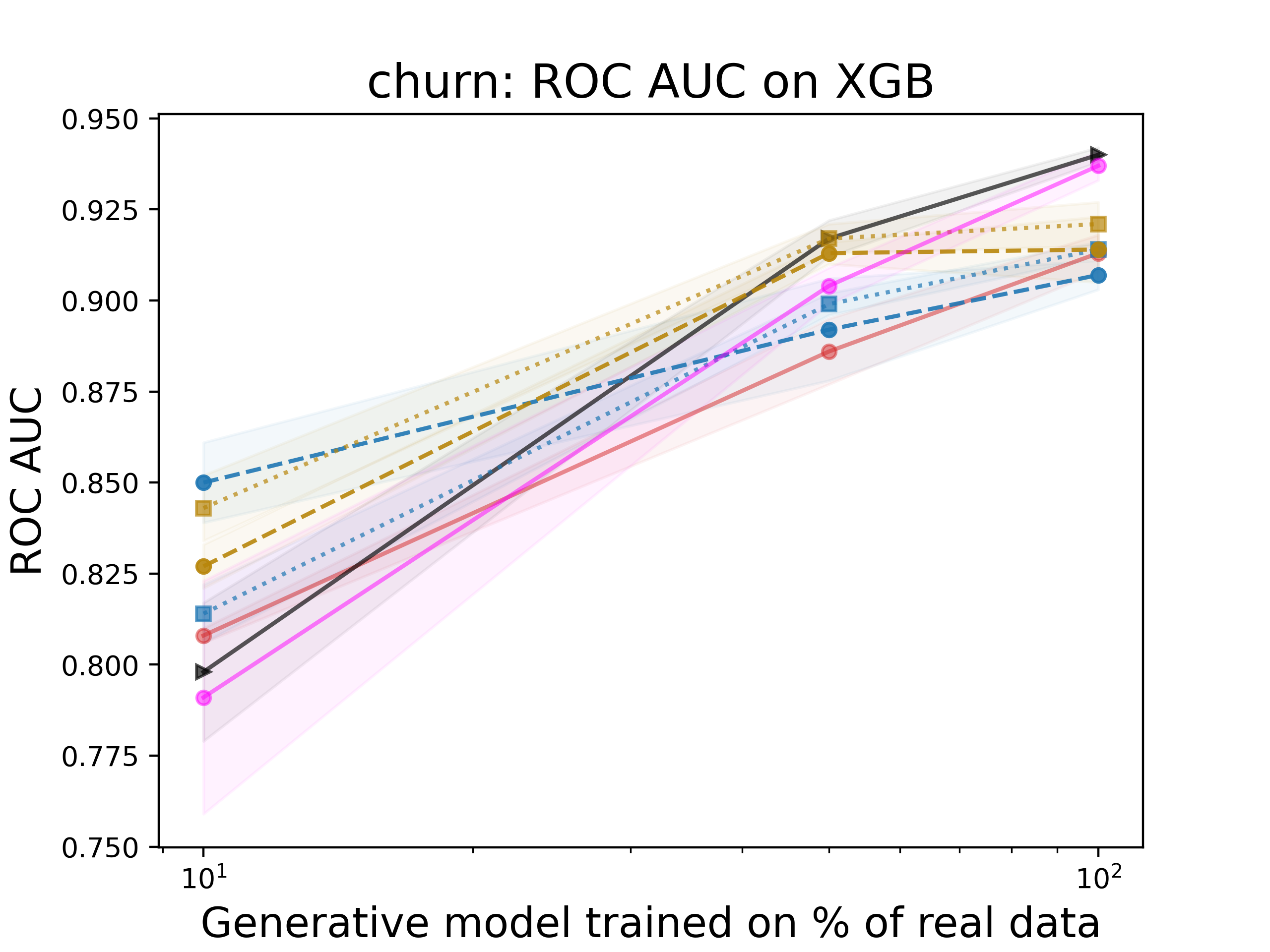}
    \end{minipage}
    \hfill
    \begin{minipage}[t]{0.245\textwidth}
        \centering
        \includegraphics[width=\linewidth]{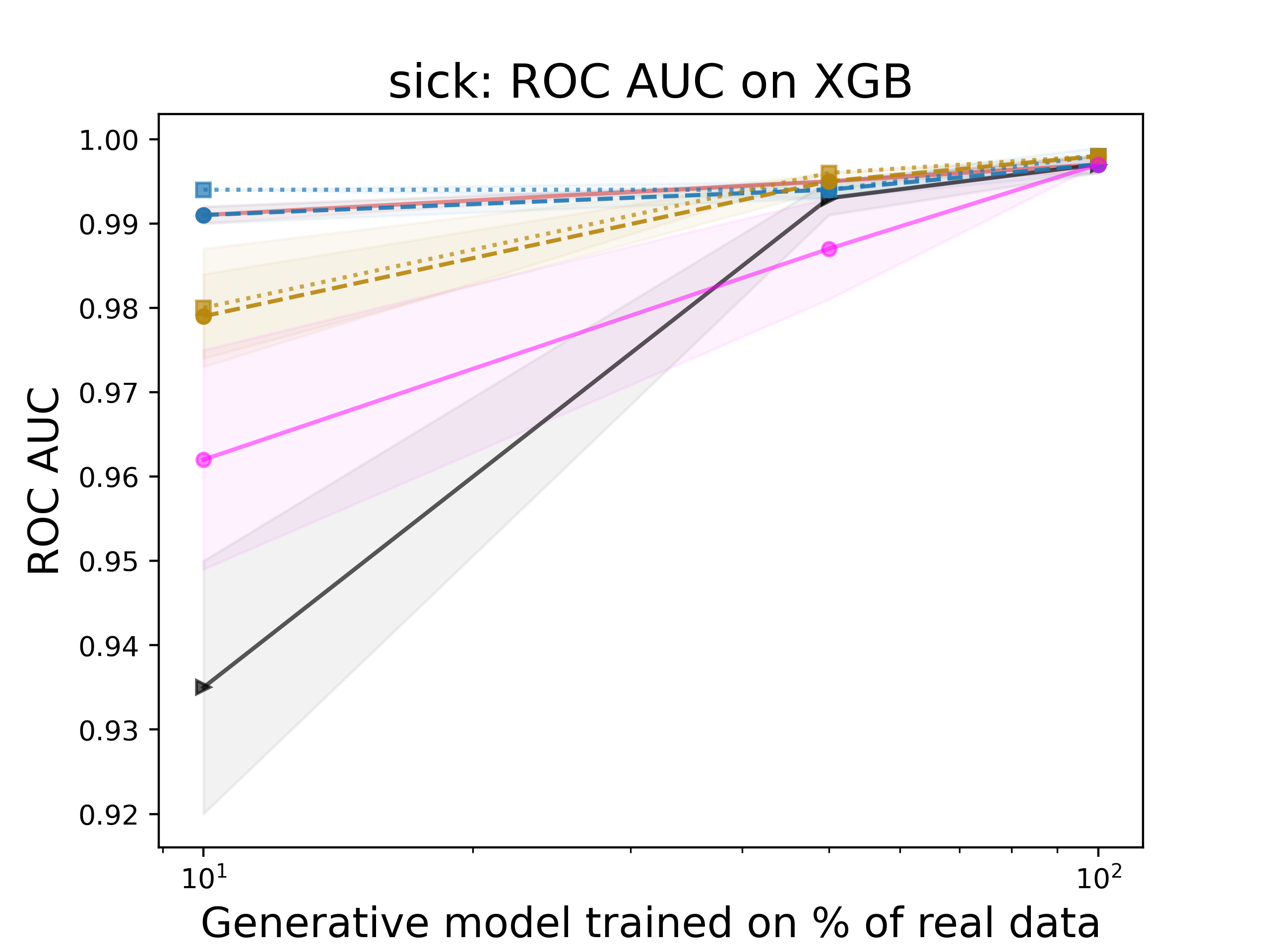}
    \end{minipage}
    \hfill
    \begin{minipage}[t]{0.245\textwidth}
        \centering
        \includegraphics[width=\linewidth]{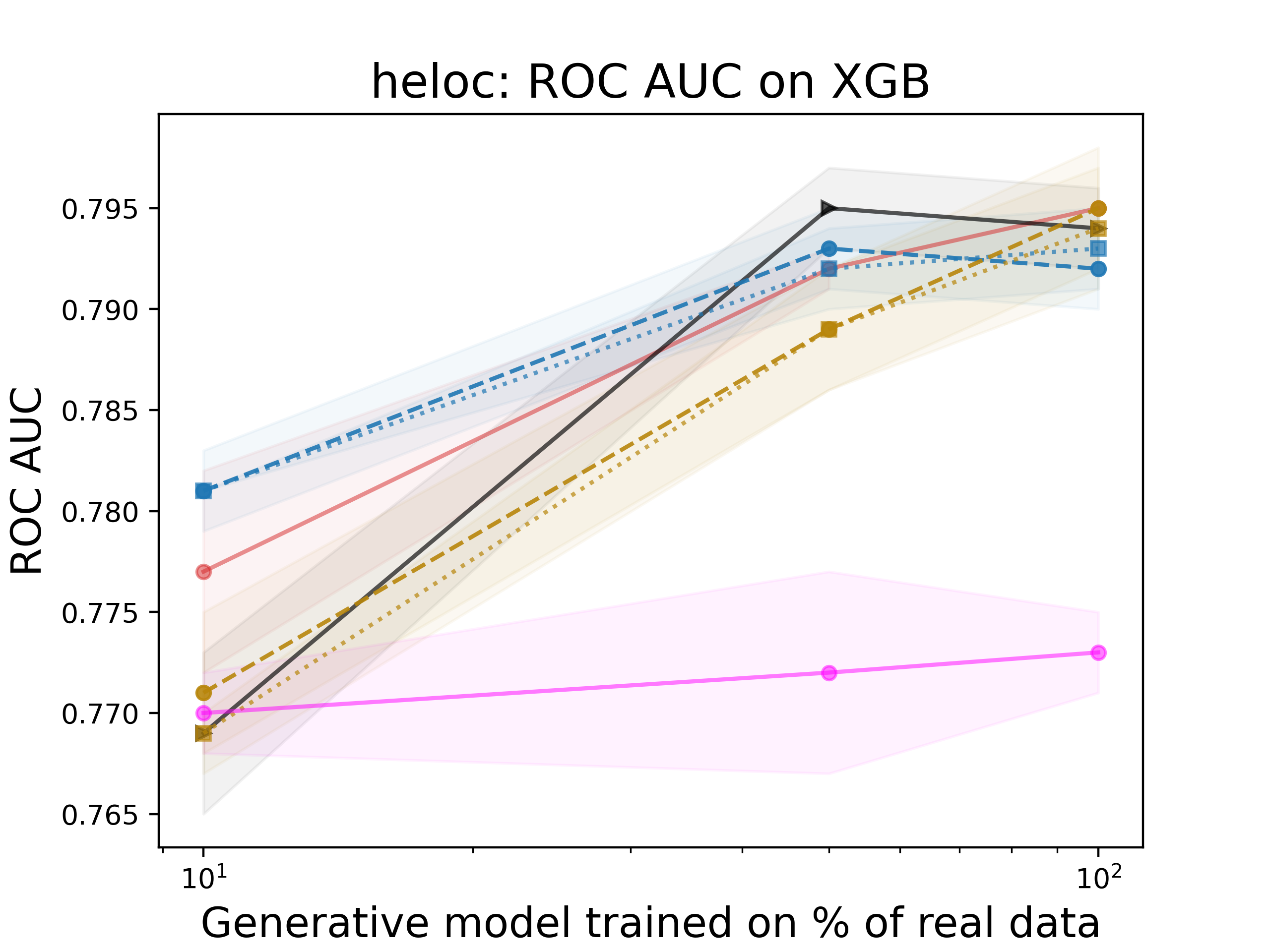}
    \end{minipage}
    \caption{(\emph{Augmentation experiment}) F1-score and ROC AUC on a downstream XGB model for all classification datasets.}
    \label{fig:appendix-class-augmentation}
\end{figure*}

\begin{figure*}[!t]
        \centering
        \begin{minipage}[t]{0.325\textwidth}
        \includegraphics[width=\linewidth]{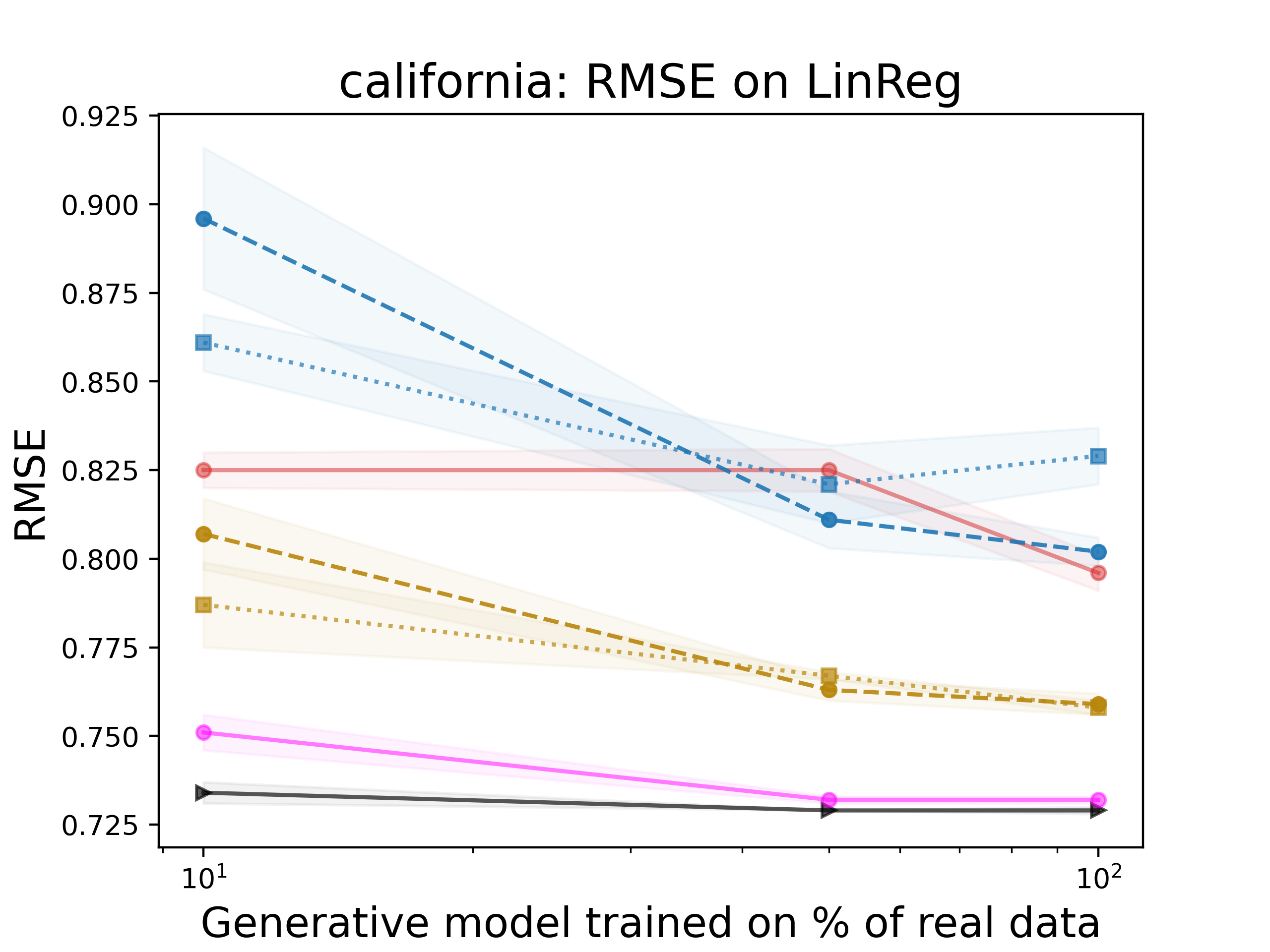}
        \end{minipage}%
        \begin{minipage}[t]{0.325\textwidth}
        \includegraphics[width=\linewidth]{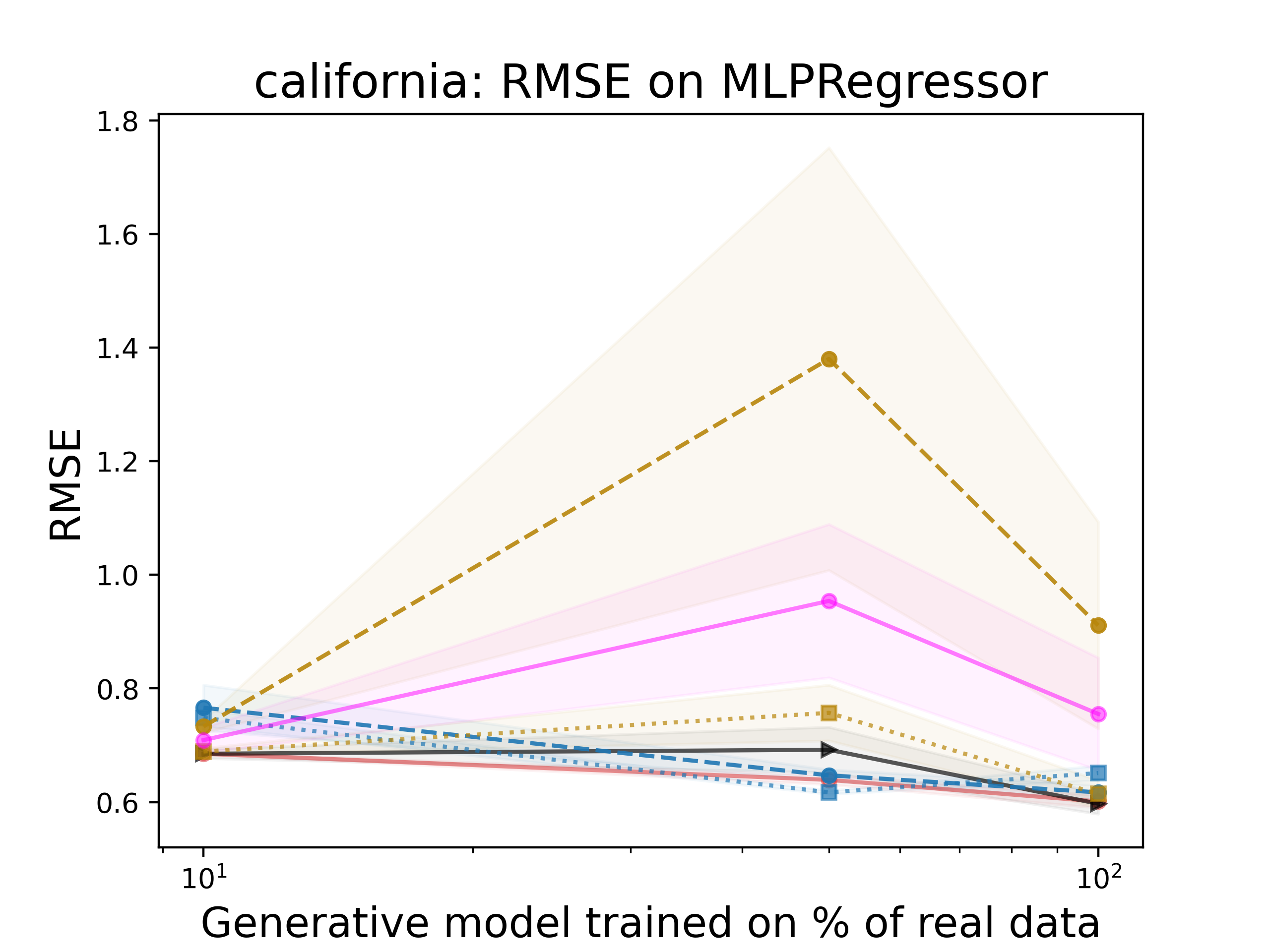}
    \end{minipage}
    \caption{(\emph{Data augmentation experiment}) Predictive metrics for downstream regression tasks  on a linear model and a multi-layered
perceptron.}
        \label{fig:appendix-regression-augmentation}
\end{figure*}

In~\cref{fig:appendix-class-augmentation,fig:appendix-class-summarization,fig:appendix-regression-augmentation,fig:appendix-summarization-regression,fig:appendix-synthetic-classification,fig:appendix-synthetic-regression} we present further predictive performances metrics and downstream models used for the evaluation of the synthetic data utility in the usecases of our experiments.

\section{Reproducibility}
\label{app:reproducibility}

Here we add some additional details on our experimentation for reproducibility. The selected hyperparameters for the trained DGMs are provided in~\cref{tab:dgm-hyperparams}. Regarding the downstream classifiers, corresponding search spaces for HPO are displayed in~\cref{tab:hpo-search-spaces}.

% Please add the following required packages to your document preamble:
% \usepackage{booktabs}
\begin{table*}[!ht]
\centering 
\begin{tabular}{@{}l||ll@{}}
\toprule
TVAE &
  \begin{tabular}[c]{@{}l@{}}n\_units\_embedding=500, \\ lr=5e-4, \\ weight\_decay=1e-5, \\ batch\_size=1000, \\ decoder\_n\_layers\_hidden=2, \\ decoder\_n\_units\_hidden=256, \\ decoder\_nonlin="leaky\_relu", \\ decoder\_dropout=0.1, \\ encoder\_n\_layers\_hidden=3, \\ encoder\_n\_units\_hidden=256, \\ encoder\_nonlin="leaky\_relu", \\ encoder\_dropout=0.1, \\ loss\_factor=1, \\ data\_encoder\_max\_clusters=10, \\ clipping\_value=1, \\ patience=500\end{tabular} &
   \\ \midrule
CTGAN &
  \begin{tabular}[c]{@{}l@{}}generator\_n\_layers\_hidden=2, \\ generator\_n\_units\_hidden=256, \\ generator\_nonlin="relu", \\ generator\_dropout=0.1, \\ generator\_opt\_betas=(0.9, 0.999), \\ discriminator\_n\_layers\_hidden=2, \\ discriminator\_n\_units\_hidden=256, \\ discriminator\_nonlin="leaky\_relu", \\ discriminator\_n\_iter=1, \\ discriminator\_dropout=0.1,\\ discriminator\_opt\_betas=(0.9, 0.999), \\ lr=5e-4, \\ weight\_decay=1e-3, \\ batch\_size=1000, \\ clipping\_value=1, \\ lambda\_gradient\_penalty=10, \\ encoder\_max\_clusters=10, \\ patience=500,\end{tabular} &
   \\ \midrule
NF &
  \begin{tabular}[c]{@{}l@{}}n\_layers\_hidden=2,\\ n\_units\_hidden=256,\\ batch\_size=1000,\\ num\_transform\_blocks=1,\\ dropout=0.1,\\ batch\_norm=False,\\ num\_bins=8,\\ tail\_bound=3,\\ lr=5e-4,\\ apply\_unconditional\_transform=True,\\ base\_distribution="standard\_normal",\\ linear\_transform\_type="permutation",\\ base\_transform\_type="rq-autoregressive",\\ encoder\_max\_clusters=10,\\ n\_iter\_min=100,\\ patience=500,\end{tabular} &
   \\ \bottomrule
\end{tabular}
\caption{Hyperparameters used for our experiments with deep generative models.}
\label{tab:dgm-hyperparams}
\end{table*}

\begin{table*}[!ht]
\centering
\begin{tabular}{>{\raggedright}p{3.5cm} p{5.5cm}}
\toprule
\textbf{XGB} & \\
\[
\begin{aligned}
    & \text{max\_depth} \in \{2, 4, 8, 12\}, \\
    & \text{learning\_rate} \in \{1e-6, 1e-4, 1e-3, 1e-2, 1e-1, 1.0\}, \\
    & \alpha \in \{0.0, 1e-6, 1e-4, 1e-2, 1e-1, 1.0\}, \\
    & \lambda \in \{1e-8, 1e-6, 1e-4, 1e-2, 1e-1, 1.0\}, \\
    & \gamma \in \{0.0, 0.1, 0.2, 0.3, 0.4, 0.5, 1.0\}, \\
    & \text{min\_child\_weight} \in \{1, 3, 5, 7, 9\}
\end{aligned}
\]
& \\
\midrule
\textbf{MLP Regressor} & \\
\[
\begin{aligned}
    & \text{hidden\_layer\_sizes} \in \{(100,), (200,)\}, \\
    & \alpha \in \{1e-4, 1e-3\}, \\
    & \text{learning\_rate} \in \{\text{"constant"}, \text{"invscaling"}\}, \\
    & \text{learning\_rate\_init} \in \{1e-3, 1e-4\}
\end{aligned}
\]
& \\
\bottomrule
\end{tabular}
\caption{Search spaces for hyperparameter optimization on downstream models.}
\label{tab:hpo-search-spaces}
\end{table*}

\end{document}